\documentclass[final,  3p, times]{elsarticle}

\usepackage{lineno}
\modulolinenumbers[5]

\usepackage{graphicx}
\usepackage{amsmath}
\usepackage{amssymb}
\usepackage{booktabs}
\usepackage{verbatim}
\usepackage{subcaption}
\usepackage{mathtools}

\usepackage{url}
\usepackage{times}
\usepackage{epsfig}
\usepackage{subcaption}
\usepackage{multirow}
\usepackage{comment}
\usepackage{balance}
\usepackage{arydshln}
\usepackage{adjustbox}

\graphicspath{ {imgs/} }

\usepackage{dsfont}
\usepackage{mathtools}
\usepackage{booktabs}
\usepackage{graphics}
\usepackage[mathscr]{euscript}

\usepackage[pagebackref,breaklinks,colorlinks]{hyperref}
\usepackage{algorithm}
\usepackage{algorithmic}
\usepackage{cleveref}
\usepackage{amsfonts}
\usepackage[normalem]{ulem}
\usepackage{xspace}
\usepackage{dsfont}
\usepackage[accsupp]{axessibility}

\usepackage{xcolor,colortbl}

\definecolor{Gray}{gray}{0.9} 

\usepackage{pifont}

\def\onedot{\ifx\@let@token.\else.\null\fi\xspace}
\def\eg{\emph{e.g}\onedot} 
\def\ie{\emph{i.e}\onedot}

\def\etal{\emph{et al}\onedot}

\makeatother

\Crefname{section}{Sec.}{Secs.}
\Crefname{section}{Section}{Sections}
\Crefname{table}{Table}{Tables}
\crefname{table}{Tab.}{Tabs.}

\usepackage{siunitx}
\usepackage{makecell}
\usepackage{boldline}

\usepackage{amsthm}
\newtheorem{theorem}{Theorem}

\definecolor{navy}{RGB}{0, 0, 128} 

\newcommand{\vx}{\mathbf{x}}
\newcommand{\vy}{\mathbf{y}}

\newcommand{\tta}{\boldsymbol{\theta}}

\newcommand{\emb}[1]{\mathbf{z}_{#1}}
\newcommand{\lgt}[1]{\mathbf{o}_{#1}}
\newcommand{\data}[1]{\mathcal{D}_{#1}}
\newcommand{\mdl}[3]{f(#1 ;\boldsymbol{\theta}_{#2}^{#3})}
\newcommand{\mpr}[1]{\boldsymbol{\theta}_{#1}}
\newcommand{\hpr}[1]{\boldsymbol{\phi}_{#1}}
\newcommand{\gpr}[1]{\boldsymbol{\psi}_{#1}}
\newcommand{\nhpr}[1]{\boldsymbol{\varphi}_{#1}}

\definecolor{mycolor}{HTML}{CCCCFF}
\definecolor{LightCyan}{rgb}{0.88,1,0.88}
\definecolor{dp}{rgb}{0, 0.0, 0}
\newcommand{\highest}[1]{\textbf{#1}}
\definecolor{mygray-bg}{gray}{0.9}

\journal{Neural Networks}

\makeatletter
\def\ps@pprintTitle{%
 \let\@oddhead\@empty
 \let\@evenhead\@empty
 \let\@oddfoot\@empty
 \let\@evenfoot\@empty
}
\makeatother
\begin{document}

\begin{frontmatter}

\title{Flashbacks to Harmonize Stability and Plasticity in Continual Learning\tnoteref{t1}}
\tnotetext[t1]{Manuscript submitted to \textit{Neural Networks (Elsevier)} in August 2024; and accepted in May 2025 for publication.}
\author[1,3]{Leila Mahmoodi\corref{c1}}
\ead{leila.mahmoodi@{monash.edu, csiro.au}}
\author[3,4]{Peyman Moghadam}
\ead{peyman.moghadam@{csiro.au, qut.edu.au}}
\author[1]{Munawar Hayat}
\ead{munawar.hayat@monash.edu}
\author[2]{Christian Simon}
\ead{christian.simon@sony.com}
\author[1,3]{Mehrtash Harandi}
\ead{mehrtash.harandi@monash.edu}

\cortext[c1]{Corresponding author; Leila Mahmoodi}

\address[1]{Monash University; Melbourne, VIC, Australia}
\address[3]{CSIRO, Data61; Brisbane, QLD, Australia}
\address[4]{Queensland University of Technology; Brisbane, QLD, Australia}
\address[2]{SONY; Japan}

\renewcommand*{\thefootnote}{\fnsymbol{footnote}}

\begin{abstract}

\begingroup
\color{dp}
We introduce Flashback Learning (FL), a novel method designed to harmonize the stability and plasticity of models in Continual Learning (CL). Unlike prior approaches that primarily focus on regularizing model updates to preserve old information while learning new concepts, FL explicitly balances this trade-off through a \textit{bidirectional} form of regularization. This approach effectively guides the model to swiftly incorporate new knowledge while actively retaining its old knowledge.
FL operates through a \textit{two-phase} training process and can be seamlessly integrated into various CL methods, including replay, parameter regularization, distillation, and dynamic architecture techniques. In designing FL, we use two distinct \textit{knowledge bases}: one to enhance plasticity and another to improve stability. FL ensures a more balanced model by utilizing both knowledge bases to regularize model updates.
Theoretically, we analyze how the FL mechanism enhances the stability-plasticity balance. Empirically, FL demonstrates tangible improvements over baseline methods within the same training budget. By integrating FL into at least one representative baseline from each CL category, we observed an average accuracy improvement of up to 4.91\% in Class-Incremental and 3.51\% in Task-Incremental settings on standard image classification benchmarks. Additionally, measurements of the stability-to-plasticity ratio confirm that FL effectively enhances this balance. FL also outperforms state-of-the-art CL methods on more challenging datasets like ImageNet. The codes of this article will be available at \href{https://github.com/csiro-robotics/Flashback-Learning}{https://github.com/csiro-robotics/Flashback-Learning}.

\endgroup
\end{abstract}


\end{frontmatter}



\section{Introduction}
\begin{figure}
\centering
\includegraphics[width=0.5\columnwidth]{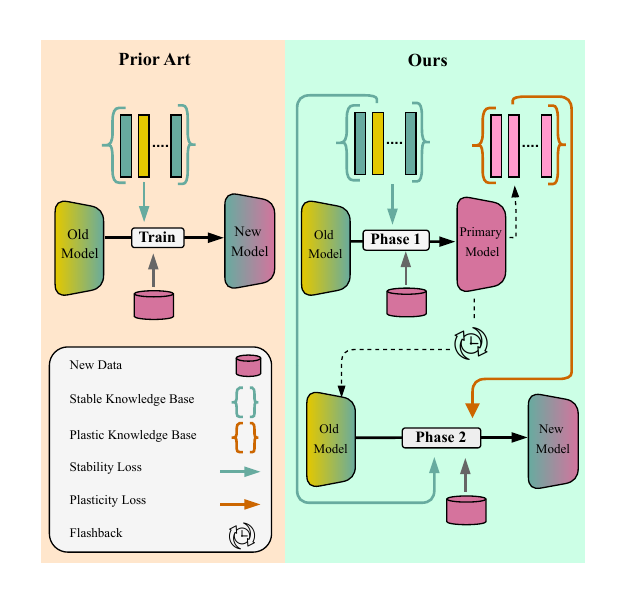} 
\caption{\textbf{Left:} Former CL methods use existing knowledge from previous tasks to adjust model updates on new data. \textbf{Right:} FL method refines new knowledge out of new task data in Phase 1, then controls model updates by bidirectional regularization in Phase 2.} 

\label{fig:summary}
\end{figure}

Our brain excels at learning new information without significantly disrupting or forgetting previously acquired knowledge. In contrast, Deep Neural Networks (DNNs) often struggle with Catastrophic Forgetting (CF)~\cite{robins1995catastrophic}, where new information can overwrite and interfere with existing knowledge. This phenomenon poses a significant challenge when continuous learning of new tasks is required. Continual Learning (CL) methods have been developed to mitigate catastrophic forgetting by enabling DNNs to retain prior knowledge while acquiring new skills.

Deploying AI models in realistic settings demands that models function effectively when encountering data with varying distributions over time. Therefore, updating the model based on newly received data after training on previous tasks is necessary. However, storing all previous data and retraining the model on the entire dataset from scratch is not feasible due to storage, processing, or privacy constraints. Consequently, many studies have focused on developing methods to learn from new data while avoiding CF. The primary challenge in this endeavor is to achieve a delicate balance between the model's \textbf{stability} (\ie, the ability to preserve what it has learned before) and its \textbf{plasticity} (\ie, the capacity to learn new data)~\cite{de2021continual}. The more effectively this balance is maintained, the more optimal the model's performance in the continual learning setting.

Towards this aim, CL methods employ various techniques, such as \textbf{1.} constraining the model parameters (\eg, \cite{chaudhry2018riemannian,nie2023bilateral,schwarz2018progress}), \textbf{2.} knowledge distillation (\eg, \cite{li2017learning},\cite{douillard2021plop},\cite{roy2023subspace}), or \textbf{3.} replaying old memory~\cite{hayes2021replay} using a fraction of old samples (\eg, \cite{bonicelli2022effectiveness,rebuffi2017icarl}) or generative models to recreate them (\eg, \cite{shin2017continual,van2020brain,mundt2022unified}). These methodologies have a common feature: they retain some knowledge from previous tasks and use it during model updates on new task data. 
We call this retained knowledge ``\textbf{stable knowledge}" in this study. ``\textbf{stable knowledge}" represents the information from previous tasks that CL methods aim to embed within the model while learning novel concepts and information. 
To successfully learn continually, stability must be balanced with plasticity. However, as revealed by new studies (\eg, \cite{kim2023stability}) most CL methods prioritize stability over plasticity. In \cite{kim2023stability}, Kim \etal evaluate several distinct CL methods in terms of their ability to extract useful features from new task data after the first tasks. They show that the model's feature extractor's capacity deteriorates significantly in most methods, while their primary focus has been on maintaining stability.
When stability is prioritized, the model essentially operates in a \textit{unidirectional} flow to transfer knowledge from old tasks to new tasks (see Figure~\ref{fig:summary}: \textbf{Left}). Excessive emphasis on stability can hinder the model's ability to learn new tasks effectively. A missing key here is that CL methods often do not have explicit control over plasticity despite model access to a new source of knowledge when a new task begins. This gap prompts a question: \emph{Once the new task begins, can we harness its knowledge to control stability-plasticity balance effectively?} Imagine compiling new task knowledge into a new source called ``\textbf{plastic knowledge}" and enabling the CL method to take control over plasticity (see \Cref{fig:summary}: \textbf{Right} for a conceptual illustration).

Achieving this would establish a \textit{bidirectional} formulation of knowledge transfer to guide the model updates while ensuring a balanced approach, harmonizing stability and plasticity effectively; thereby, stability is not prioritized over plasticity. \textcolor{dp}{Interestingly, evidence from neuroscience supports the feasibility of such a bidirectional mechanism. The stability of dendritic spines in the neocortex is associated with lifelong consolidated memories~\cite{yang2009stably}, and the hippocampus plays a crucial role in memory consolidation by coordinating memory replay with the visual cortex during sleep~\cite{ji2007coordinated}. ~\cite{ji2007coordinated} shows multicell firing sequences, evoked by awake experiences, are replayed in both the cortex and hippocampus during sleep, leading to the simultaneous reactivation of coherent memory traces in both regions. This suggests that the biological replay mechanism is not merely unidirectional from the hippocampus to the neocortex. The hippocampus actively refines memories by integrating new information into past memory traces~\cite{gonzalez2020can}. This process fosters a more profound understanding and better adaptation to the new context~\cite{bridge2014hippocampal}.} 

In this study, we build upon our initial research \cite{mahmoodi2023flashback} on the impact of bidirectional knowledge flow on the stability-plasticity balance. The encouraging results from that investigation motivated us to develop a mechanism that can be adapted to various categories of CL methods. We demonstrate the benefits of this approach across a broad range of CL techniques. To be more specific, we formulate FL for distillation, memory rehearsal, parameter regularization, and dynamic architecture methods. The first phase of FL focuses on compiling plastic knowledge as the new task commences and its data becomes available. We update the old model with new task data throughout this phase, prioritizing the swift acquisition of new information over preserving the old one. This phase creates a primary version of the new model, which is essential for accumulating plastic knowledge. 
In the second phase, termed ``flashback," the old model revisits new task data once more. This phase guides model updates through a bidirectional knowledge flow, in which one side utilizes stable knowledge to enforce stability while the other employs plastic knowledge to drive plasticity. \textcolor{dp}{Theoretically, we show that the bidirectional knowledge flow moves the model weights towards maintaining an interpolation between old and new tasks' representations, directly contributing to stability-plasticity balance.}

We conduct comprehensive experiments to demonstrate the effectiveness of the FL mechanism in improving CL methods. Our evaluation covers several established CL image classification benchmarks, including Split CIFAR-10, Split CIFAR-100, and Split Tiny ImageNet, across two primary settings: Class-Incremental (CI) and Task-Incremental (TI) Learning. 
We assess at least one method from each CL category on all benchmarks under both CI and TI settings, including LUCIR~\cite{hou2019learning} and LwF~\cite{li2017learning} from distillation methods, iCaRL~\cite{rebuffi2017icarl} and X-DER~\cite{boschini2022class} from replay-based methods, oEWC~\cite{schwarz2018progress} from parameter regularization approaches, and FOSTER~\cite{wang2022foster} and BEEF~\cite{wang2023beef} from architecture-based methods. Our evaluation reports standard CL metrics computed before and after integrating the FL mechanism with each CL method. We observe FL significantly improves CL methods' average accuracy up to 4.9\% and 3.5\%, respectively, under CI and TI settings within the same training budget. Moreover, FL effectively reduces average forgetting and preserves performance on old tasks in CL methods. In summary, this work makes the following contributions:

\begin{itemize}
    \item We propose flashback learning, providing a mechanism for existing CL methods to improve their stability-plasticity balance. FL functions in two phases: building an expert knowledge base from new task data to promote plasticity and utilizing it to counteract the unidirectional stability regularization in common CL methods. 
    \item \textcolor{dp}{We theoretically show how bidirectional regularization in the second phase of our algorithm enhances the stability-plasticity balance in previous CL methods.}
 
    \item We develop and demonstrate the efficacy of FL across four main categories of CL methods: 1) parameter constraint-based methods, 2) knowledge distillation methods, 3) replay memory methods (including both partial old sample reuse and generative memory), and 4) architecture expansion methods. Our experiments show that FL provides tangible improvements in all these categories.
    \item We conduct comprehensive experiments to benchmark FL against several distinct CL methods across multiple datasets. Our results consistently indicate that FL enhances performance, demonstrating substantial improvements even when integrated with advanced CL strategies.
\end{itemize}

\section{Notations}
\label{sec:method}

In the CL setting, a model is trained sequentially on a series of tasks, denoted by $\mathbb{T} = \{\mathcal{T}_t\}_{t=1}^T$. For each task \(\mathcal{T}_t\), the inputs $\vx \in \mathcal{X}_t$ and their ground truth labels $\mathbf{y} \in \mathcal{Y}_t$ are drawn from a data distribution $\mathcal{D}_t$. Here, $\mathcal{X}_t$ and $\mathcal{Y}_t$ represent the task-specific sets of inputs and outputs, respectively. Data in task $\mathcal{T}_t$ originated from a distinct set of classes $C_t$, with no overlap between classes across different tasks. Task identifier $t$ is available during evaluation in the Task-Incremental (TI) setting, allowing the model to concentrate decision boundaries within individual tasks. In the Class-Incremental (CI) setting, the model must decide on all seen classes during inference.
\begingroup
\color{dp}
 Throughout this paper, all vectors are represented with bold symbols (e.g., $\mathbf{v}$), and the $i^\text{th}$ element of a vector is denoted by $[i]$ following the symbol (e.g., $\mathbf{v}[i]$). We denote a model by $f(\cdot;\boldsymbol{\theta})$, where $\boldsymbol{\theta}\in\mathbb{R}^P$ is the parameter vector. The model is decomposed into two parts: feature extractor $h(\cdot;\boldsymbol{\phi})$ and classifier $g(\cdot;\boldsymbol{\psi})$, \, ie $f(\cdot;\boldsymbol{\theta})=g(h(\cdot;\,\boldsymbol{\phi});\boldsymbol{\psi})$ and $\boldsymbol{\theta}=\{\boldsymbol{\phi},\boldsymbol{\psi}\}$ where $\boldsymbol{\phi}$ and $\boldsymbol{\psi}$ are the parameter vectors of feature extractor and classifier, respectively. While training on task $\mathcal{T}_t$, we show all parameters subscript with $t$ (\eg, $\mpr{t}$), and after task $\mathcal{T}_t$, we show the optimal parameters with $\boldsymbol{*}$ superscript (\eg, $\mpr{t}^*$), see \ref{sec:not-table} for detailed notation. Feature embeddings and classification logits are denoted by $\emb{} \in \mathbb{R}^d$ and $\lgt{} \in \mathbb{R}^{C_t}$; $d$ is the dimension of feature embedding space, and $C_t$ is the total number of classes seen till task~$\mathcal{T}_t$. We apply $t$ subscript for feature embeddings and logits of task~$\mathcal{T}_t$ (\eg, $\emb{t} = h(\vx;\hpr{t})$ and $\lgt{t}=\mdl{\vx}{t}{}$).
 \endgroup 

\section{Flashback Learning}
\begin{figure}
\centering
\includegraphics[width=\textwidth]{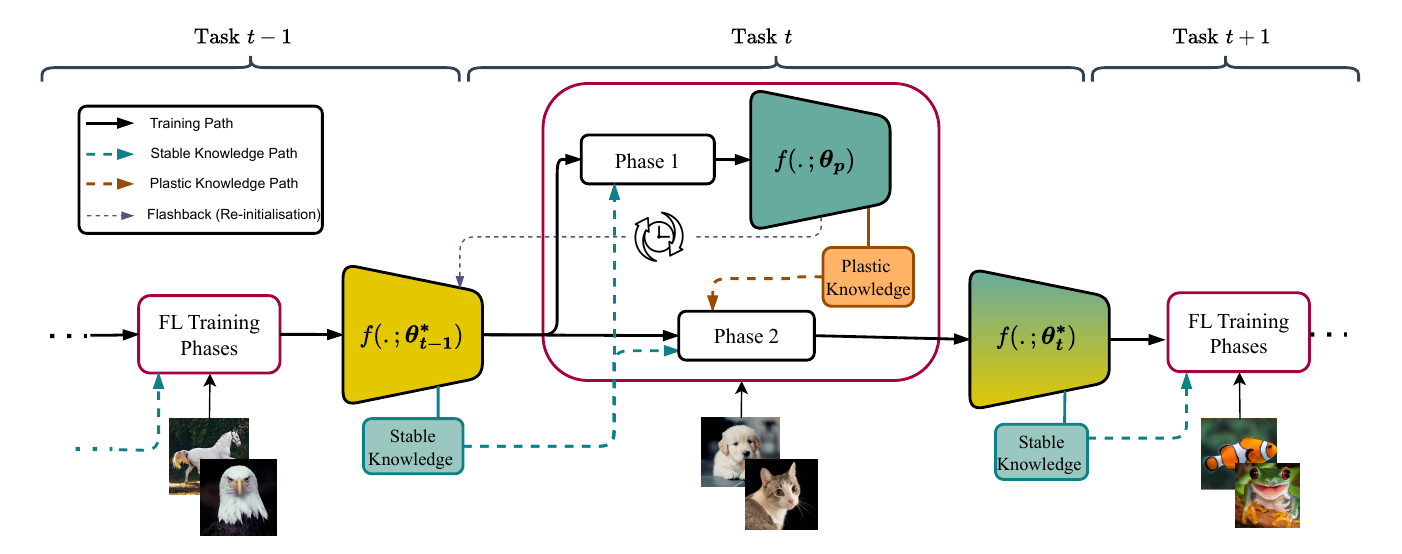}
\caption{Flashback Learning overview; At task~$\mathcal{T}_t$, \textbf{Phase~1}- updates the old model on new data to obtain primary model $\mdl{\cdot}{p}{}$. Then, it extracts new knowledge from the primary model and stores it in PKB. \textbf{Phase~2}-flashbacks from primary to the old model $\mdl{\cdot}{t-1}{*}$; while using stable knowledge and plastic knowledge to regularize model updates in a bidirectional flow and obtain new model $\mdl{\cdot}{t}{*}$.}
\label{fig:method}
\end{figure}

CL methods update the model sequentially on tasks $\mathbb{T} = \{\mathcal{T}_t\}_{t=1}^T$; within this sequential update, at each task $\mathcal{T}_t$, the model should ideally learn new information from task $\mathcal{T}_t$ and keep what it has learned from previous tasks $\{\mathcal{T}_i\}_{i=1}^{t-1}$. Each method stores some information from previous tasks—we denote it as  \textbf{stable knowledge} in this study — and develops a strategy to maintain this information within the model while updating on task $\mathcal{T}_t$. Stable knowledge can range from a copy of the old model to a memory of past samples, and the corresponding strategy varies from distillation to replay accordingly. However, the fundamental pattern is the same - stable knowledge is transferred from previous tasks to the new model, and the main focus is on model stability by enforcing previously acquired knowledge to remain within the model throughout updates on task $\mathcal{T}_t$. While success for CL models relies on balancing stability and plasticity, overemphasizing one side can compromise the other. Most current strategies in CL methods dominantly focus on maintaining stability during updates on the new task, which can consequently hinder the model's plasticity, and the model cannot learn effectively from new data. Therefore, it is crucial to give adequate emphasis to both stability and plasticity in the model. To this end, we introduce \textbf{Flashback Learning (FL)}, a two-phase plugin mechanism compatible with various CL methods.

FL formulates a bidirectional knowledge transfer to concurrently promote model stability and plasticity, improving their balance in CL methods. When integrating FL with CL methods, we aim to refine some knowledge from task $\mathcal{T}_t$ to create a new source promoting plasticity, named ``\textbf{plastic knowledge}". Then, stable and plastic knowledge makes bidirectional knowledge transfer feasible in a subsequent step. In this bidirectional transfer, one side works towards stability and the other towards plasticity, ultimately achieving a balance. FL operates in \underline{\textit{two phases}}: \textit{Phase~1} is designed to extract plastic knowledge from task $\mathcal{T}_t$ by which we control plasticity in the next phase, and \textit{Phase~2} aims to harmonize stability-plasticity balance by leveraging stable and plastic knowledge. When task $\mathcal{T}_t$ commences, we define \textbf{stable model} by $\mdl{\cdot}{t-1}{*}$— model learned in prior tasks— and denote stable model by $\mdl{\cdot}{s}{}$ in FL algorithm, where $\mpr{s}=\mpr{t-1}^*$. 

During \underline{\textit{Phase~1}} (see Figure~\ref{fig:method}), we start from $\mpr{t-1}^*$ and update the model on $\mathcal{D}_t$ to obtain a primary version of the new model, denoted \textbf{primary model} $\mdl{\cdot}{p}{}$, here $\mpr{p}$ shows the model parameters after seeing $\mathcal{D}_t$ in \textit{Phase~1}. This initial phase prioritizes the rapid acquisition of task $\mathcal{T}_t$ information, deferring the incorporation of previously learned knowledge. Consequently, the number of training iterations is limited, and the transfer of stable knowledge to the new model is not pursued at this stage. Different categories of CL methods can act as a host for the FL algorithm. These categories include \textbf{1.} distillation, \textbf{2.} memory replay, \textbf{3.} parameter regularization, and \textbf{4.} dynamic architecture methods. The stable knowledge that a host method carries from past tasks depends on its strategy for maintaining stability. For example, a distillation-based method has access to a copy of the old model during task $\mathcal{T}_t$, which distills knowledge into the new model and preserves stability. In FL, we keep the stable knowledge carried by the host method in the \textbf{Stable Knowledge Base (SKB)}. We identify what information a host method from different CL categories holds after task $\mathcal{T}_{t-1}$ for its stability retention strategy in task $\mathcal{T}_t$ and store the same information in SKB without incurring additional expenses (see \textsection~\ref{sec:skb}.)

In \textit{Phase~1}, primary model $\mdl{\cdot}{p}{}$ gains new insights of task $\mathcal{T}_t$. This model plays the main role in compiling plastic knowledge. We ensure that plastic knowledge is compatible with the stable knowledge that the host CL algorithm has retained from the past. We store this plastic knowledge in the \textbf{Plastic Knowledge Base (PKB)}. PKB may vary within different CL methods to maintain consistency with stable knowledge. For instance, distillation-based methods hold a copy of the old model; accordingly, we keep a copy of the primary model in PKB. We provide further details on how to derive plastic knowledge for each CL category in \textsection~\ref{sec:pkb}.

\underline{\textit{Phase~2}} (see Figure~\ref{fig:method}) involves a flashback to the starting point, which means resetting the model to the stable model ($\mdl{\cdot}{s}{}$) and revisiting $\data{t}$. FL aims to balance stability and plasticity in this phase through a bidirectional regularization process. FL establishes a two-way flow of knowledge by using the information compiled in the former phase's SKB and PKB to regularize the model. The bidirectional regularization process includes two loss components. The first component uses knowledge from SKB to encourage model stability, while the second component enhances model plasticity by utilizing knowledge from PKB to learn new tasks effectively. For more information on the FL phases, please refer to \textsection~\ref{sec:fl-steps}.

\subsection{Stable Knowledge Base}

\label{sec:skb}
Knowledge retained from the past depends on the specific strategy used by the CL method for knowledge retention. Nevertheless, its purpose remains the same across all CL models: to regulate model updates on new data and prevent forgetting. In this part, we want to identify the stable knowledge used in each CL method to avoid forgetting. We denote \textbf{stable knowledge} by $\mathbf{\mathcal{S}}$ and define it within four categories of CL methods. We need a clear definition of available $\mathcal{S}$ in the original CL method to properly extract corresponding plastic knowledge at the end of \textit{Phase~1}.

\subsubsection{Distillation}
\label{subsec:distill_cl}

\begin{figure}[h]
\centering
\includegraphics[width=0.9\textwidth]{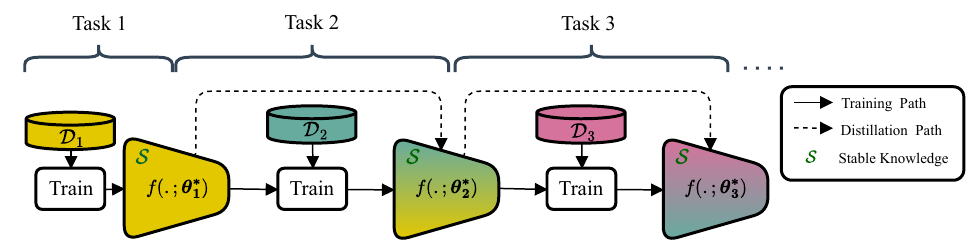}
\caption{Distillation methods keep $\mdl{\cdot}{t-1}{*}$ at end of task $t-1$ as stable knowledge for task $\mathcal{T}_t$.}
\label{fig:skb-dist}
\end{figure}

Stable model $\mdl{\cdot}{s}{}$ encapsulates the knowledge learned from previous tasks (see Figure~\ref{fig:skb-dist}); distillation methods keep a copy of this model throughout task $\mathcal{T}_t$; from which they distill into the new model at different stages— after classifier \cite{li2017learning}, in intermediate stages~\cite{douillard2020podnet}, or after feature extraction~\cite{hou2019learning}— to ensure new task representations do not deviate significantly from what the model has learned before. Here, $\mathcal{S}$ at task $\mathcal{T}_t$ is the stable model kept from prior tasks:  
\begin{equation}
\label{eq:skb-distill}
    \mathcal{S} \stackrel{\text{def}}{=} \mdl{\cdot}{s}{}= \mdl{\cdot}{t-1}{*}.
\end{equation}

\subsubsection{Memory Replay}
\label{subsec:replay_cl}

\begin{figure}[h]
\centering
\includegraphics[width=\textwidth]{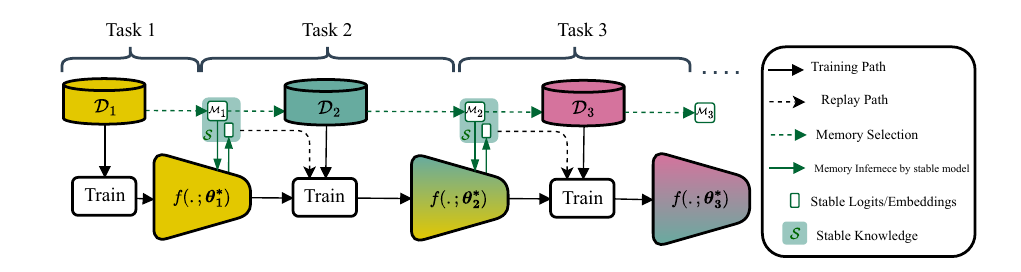}
\caption{Replay methods keep a small set of old samples $(\vx,\vy) \in \{\mathcal{D}_i\}_{i=1}^{t-1}$ in memory $\mathcal{M}_{t-1}$ at end of task $t-1$. Task $\mathcal{T}_t$'s stable knowledge includes memory samples $\mathcal{M}_{t-1}$ associated with their logits $\lgt{s}$ or embeddings $\emb{s}$.}
\label{fig:skb-replay}
\end{figure}

When task $\mathcal{T}_t$ starts, replay methods update the model on a joint distribution of new data $\mathcal{D}_t$ and limited samples held from previous tasks $\{\mathcal{T}_i\}_{i=1}^{t-1}$. In fact, a memory $\mathcal{M}_{t-1}$ with limited capacity $\lvert \mathcal{M}_{t-1}\rvert \leq M \ll \sum_{i=1}^{t-1} \lvert \mathcal{D}_i \rvert$ of past samples $(\vx,\vy) \in \{\mathcal{D}_i\}_{i=1}^{t-1}$, with their corresponding feature embeddings $\emb{}$ or logits $\lgt{}$ is the stable knowledge retained by replay methods to prevent the model from forgetting what it has acquired before (see Figure~\ref{fig:skb-replay}). For example, in DER/DER++~\cite{buzzega2020dark}, memory samples $(\vx,\vy) \in \mathcal{M}_{t-1}$ and their logits $ \lgt{} = \mdl{\vx}{s}{}$ are randomly selected and kept from previous tasks; let denote their logits by $\lgt{s}$. When updating on task $\mathcal{T}_t$, new model responses to memory samples are regularized by their available $\lgt{s}$. \cite{iscen2020memory} keeps the feature embeddings $\emb{} = h(\vx;\hpr{s})$ of memory samples for further regularization. Denote the feature embeddings by $\emb{s}$. We define $\mathcal{S}$ as selected old samples in memory $\mathcal{M}_{t-1}$ associated with their logits $\lgt{s}$ or embeddings $\emb{s}$:
\begin{equation}
\label{eq:skb-replay}
\mathcal{S} \stackrel{\text{def}}{=} \left\{ \mathbf{x}, \mathbf{y}, [\lgt{s}, \emb{s}] \mid \lgt{s}=\mdl{\vx}{s}{} ,\, \emb{s}=h(\vx;\hpr{s}) , \, \vx \in \mathcal{M}_{t-1} \right\}.
\end{equation}

\subsubsection{Parameter Regularization}
\label{subsec:param_reg_cl}

Regularization methods often assume past tasks' distributions are captured in the Fisher Information Matrices~(FIMs) calculated on previous tasks. This assumption leads to different approaches to approximate $\mathbf{F}_i$. The FIM can be used to regularize the training objective as:
\begin{equation}
\label{eq:loglike} 
    \sum_{i=1}^{t-1} \lambda_i \mathcal{L}_i \left(\tta, \mathcal{D}_i)\right) \approx \sum_{i=1}^{t-1}\lambda_i \left(\tta - {\tta_i}^*\right)^\top \mathbf{F}_i \left(\tta - {\tta_i}^*\right).
\end{equation}

Here, each component $\mathcal{L}_i \left(\tta, \mathcal{D}_i)\right)$ represents the negative log-likelihood $-\mathbb{E}_{(\vx, \vy) \sim \mathcal{D}_i} \big( - \log p\left(\vy | \vx; \tta\right)\big)$ and $\lambda_i$ regulates the importance of the i$^{th}$ task among the seen tasks. EWC \cite{kirkpatrick2017overcoming} estimates FIM of task~$\mathcal{T}_i$ by:
\begin{equation}
\label{eq:fisher}
    \mathbf{F}_i \approx \mathbb{E}_{(\vx, \vy) \sim \mathcal{D}_i} \bigg[ \Big( \nabla_{\tta} \log p (\vy | \vx; \tta ) \Big|_{\tta = \tta_i^*} \Big) ^\top \Big( \nabla_{\tta} \log p(\vy | \vx; \tta) \Big|_{\tta = \tta_i^*} \Big) \bigg]  \quad i=\{1,\dots, t-1\},
\end{equation}

where $p\left(\vy | \vx; \tta\right)$ is the probability distribution parameterized by $\tta$ and its gradient is evaluated at $\{{\tta_i}^*\}_{i=1}^{t-1}$.

\begin{figure}[h]
\centering
\includegraphics[width=\textwidth]{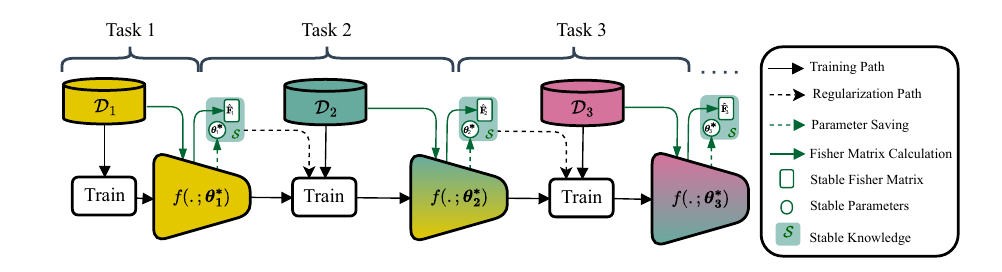}
\caption{Parameter regularization methods keep model parameters $\tta^*_{t-1}$ and their importance $\hat{\mathbf{F}}_{t-1}$ as stable knowledge for task $\mathcal{T}_t$.}
\label{fig:skb-reg}
\end{figure}

Keeping one FIM per task imposes extra complexity.  The online EWC approach \cite{schwarz2018progress} proposes an average FIM with a recursive update: 
\begin{equation}
\label{eq:fisher2}
\hat{\mathbf{F}}_{i} = \gamma \hat{\mathbf{F}}_{i-1} + \mathbf{F}_{i},
\end{equation}
where $\hat{\mathbf{F}}_i$ is the average FIM, and $\gamma<1$ is a hyperparameter to adjust the contribution of the former task's average FIM, $\hat{\mathbf{F}}_{i-1}$. Thus, for task~$\mathcal{T}_t$ the average FIM, $\hat{\mathbf{F}}_{t-1}$, is retained from the past instead of $t-1$ number of FIMs (\ie $\{\mathbf{F}_i\}_{i=1}^{t-1}$). We define stable knowledge for parameter regularization methods (see Figure~\ref{fig:skb-reg}) as:

\begin{equation}
\label{eq:skb-reg}
    \mathcal{S} \stackrel{\text{def}}{=} \{ \mpr{s} , \, \mathbf{F}_{s} \} = \{ \mpr{t-1}^* , \, \hat{\mathbf{F}}_{t-1} \}.
\end{equation}

\subsubsection{Dynamic Architecture}
\label{subsec:dynamic_arch_cl}

\begin{figure}[h]
\centering
\includegraphics[width=\textwidth]{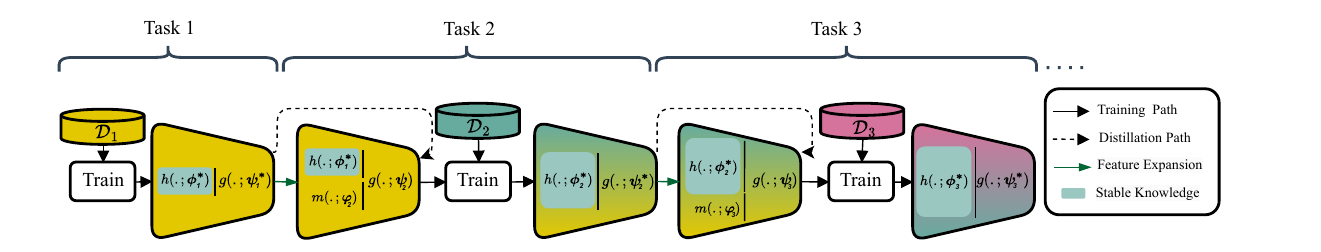}
\caption{Dynamic architecture methods keep $h(\cdot;\hpr{t-1}^*)$ at the end of task $t-1$ as stable knowledge for task $\mathcal{T}_t$.}
\label{fig:skb-dyn}
\end{figure}

Architecture-based methods commonly follow two steps: expansion (adding capacity to learn the new task) and integration (making new representations compatible with previous tasks). While the expansion is almost similar among methods in this category, the integration step is specified per method. At task~$\mathcal{T}_t$, feature extractor $h(\cdot;\hpr{t})$ built by expanding the old feature extractor $h(\cdot;\hpr{t-1}^*)$ with a new module $m(\cdot;\nhpr{t})$ \cite{yan2021dynamically} as:
\begin{equation}
\label{eq:dyn-expansion}
  h(\cdot;\hpr{t}) = \left[ h(\cdot;\hpr{t-1}^*)\,\|\, m(\cdot;\nhpr{t}) \right];
\end{equation}
the operator  $\|$  shows concatenation between $h(\cdot;\hpr{t-1}^*)$'s and $m(\cdot;\nhpr{t})$'s outputs. Adding $m(\cdot;\nhpr{t})$ provides extra plasticity for the model to learn new representations. In expansion stage, the model learns task $\mathcal{T}_t$ data while the old feature extractor $h(\cdot;\hpr{t-1}^*)$ is frozen, and only $m(\cdot;\nhpr{t})$ is updated. Then, in integration \cite{wang2022foster},\cite{yan2021dynamically}, the model is trained again to effectively make newly-learned representations by $m(\cdot;\nhpr{t})$ compatible with old ones featured by $h(\cdot;\hpr{t-1}^*)$. Hence, we define stable knowledge for dynamic architecture methods as (see Figure~\ref{fig:skb-dyn}): 
\begin{equation}
\label{eq:skb-dyn}
     \mathcal{S}  \stackrel{\text{def}}{=} h(\cdot;\hpr{s}) = h(\cdot;\hpr{t-1}^*).
\end{equation}

\subsection{Plastic Knowledge Base}
\label{sec:pkb}
Having defined stable knowledge per category in \textsection~\ref{sec:skb}, we explain how we extract plastic knowledge from new data in task $\mathcal{T}_t$. As mentioned before, FL \textit{Phase~1} updates the model on $\mathcal{D}_t$ with a focus on rapidly capturing task $\mathcal{T}_t$'s information, which results in the primary model $\mdl{\cdot}{p}{}$. At the end of \textit{Phase~1}, we use primary model $\mdl{\cdot}{p}{}$ to extract plastic knowledge from task $\mathcal{T}_t$. We denote \textbf{plastic knowledge} by $\mathcal{P}$ and define it per CL category. In doing so, we should ensure that plastic knowledge is compatible with the stable knowledge the host method holds from previous tasks. This compatibility improves equal two-way knowledge transfer from  SKB and PKB in FL \textit{Phase~2} and stability-plasticity balance.

\subsubsection{Distillation}
\label{subsec:distill_fl}

\begin{figure}[h]
\centering
\includegraphics[width=0.9\textwidth]{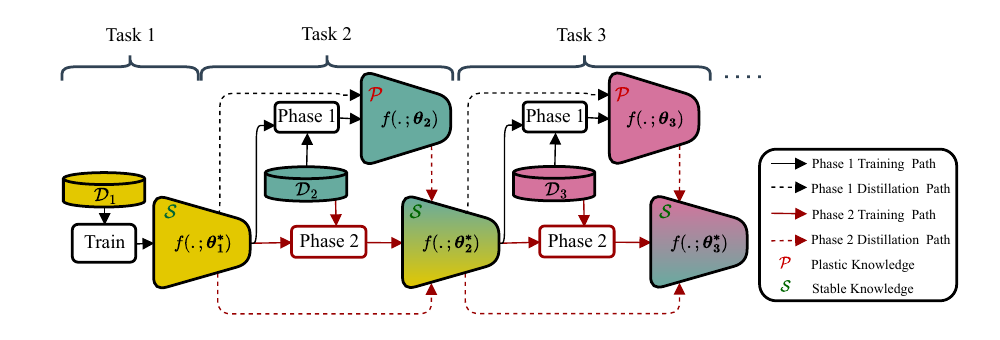}
\caption{ \textit{Phase~1} concludes by keeping $\mdl{\cdot}{t}{}$ as plastic knowledge for task $\mathcal{T}_t$.}
\label{fig:pkb-dist}
\end{figure}

Having learned the representation of previous tasks $\{\mathcal{D}_i\}_{i=1}^{t-1}$, old model $\mdl{\cdot}{t-1}{*}$~\eqref{eq:skb-distill} is held in SKB for distillation methods. We equally define plastic knowledge by the model primary model $\mdl{\cdot}{p}{}$ having seen $\mathcal{D}_t$ through \textit{Phase~1}(see Figure~\ref{fig:pkb-dist}):
\begin{equation}
\label{eq:pkb-distill}
    \mathcal{P}  \stackrel{\text{def}}{=}  \mdl{\cdot}{p}{} = \mdl{\cdot}{t}{}. 
\end{equation}

\subsubsection{Memory Replay}
\label{subsec:replay_fl}

\begin{figure}[h]
\centering
\includegraphics[width=0.95\textwidth]{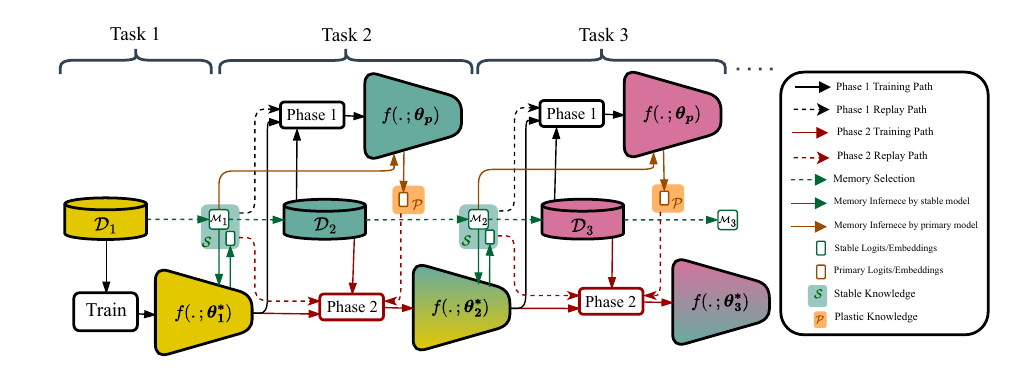}
\caption{ \textit{Phase~1} evaluates memory samples by primary model to keep $\lgt{p}$ or $\emb{p}$ as plastic knowledge for task $\mathcal{T}_t$.}
\label{fig:pkb-replay}
\end{figure}

Compatible with old logits $\lgt{s}$ or feature embeddings $\emb{s}$ kept in replay method SKB~\eqref{eq:skb-replay}, we evaluate memory samples $\vx \in \mathcal{S}$ by the primary model to obtain new logits $\lgt{p}=\mdl{\vx}{p}{}$ or feature embeddings $\emb{p}=h(x;\hpr{p})$. We keep $\lgt{p}$ or $\emb{p}$ in PKB (see Figure~\ref{fig:pkb-replay}):
\begin{equation}
\label{eq:pkb-replay}
    \mathcal{P} \stackrel{\text{def}}{=} \left\{  [\lgt{p}, \emb{p}] \mid \lgt{p}=\mdl{\vx}{p}{}\, , \emb{p}=h(\vx;\hpr{p}) \, , \vx \in \mathcal{M}_{t-1} \right\}.
\end{equation}

\subsubsection{Parameter Regularization}
\label{subsec:param_reg_fl}

\begin{figure}[h]
\centering
\includegraphics[width=\textwidth]{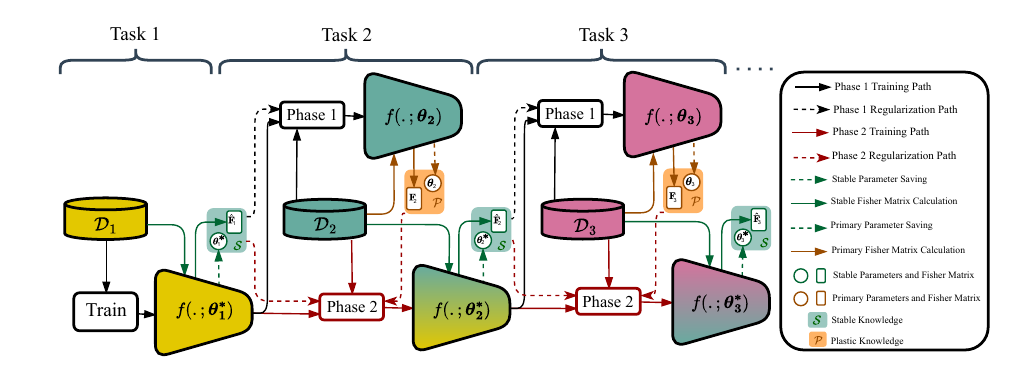}
\caption{ \textit{Phase~1} concludes by calculating $\mathbf{F}_t$ and taking $\{ \tta_t,\mathbf{F}_t\}$ as plastic knowledge for task $\mathcal{T}_t$.}
\label{fig:pkb-reg}
\end{figure}

To keep consistency with stable knowledge~\eqref{eq:skb-reg} defined in this category, we require model parameters and their FIM that have captured task~$\mathcal{T}_t$'s distribution. Having seen $\mathcal{D}_t$ through \textit{Phase~1}, model parameters $\mpr{t}$ have absorbed task~$\mathcal{T}_t$'s knowledge. We estimate task~$\mathcal{T}_t$'s FIM at the end of \textit{Phase~1} by:
\begin{equation}
\label{eq:fisher_p}
    \mathbf{F}_t \approx \mathbb{E}_{(\vx, \vy) \sim \mathcal{D}_t} \bigg[ \Big( \nabla_{\tta} \log p (\vy | \vx; \tta ) \Big|_{\tta = \tta_t} \Big) ^\top \Big( \nabla_{\tta} \log p(\vy | \vx; \tta) \Big|_{\tta = \tta_t} \Big) \bigg],
\end{equation}
and define the plastic knowledge (see Figure~\ref{fig:pkb-reg}) as:   
\begin{equation}
\label{eq:pkb-reg}
    \mathcal{P} \stackrel{\text{def}}{=} \{ \tta_p,\mathbf{F}_p\}=\{ \tta_t,\mathbf{F}_t\}.
\end{equation}

\subsubsection{Dynamic Architecture}
\label{subsec:dynamic_arch_fl}
\begin{figure}[h]
\centering
\includegraphics[width=\textwidth]{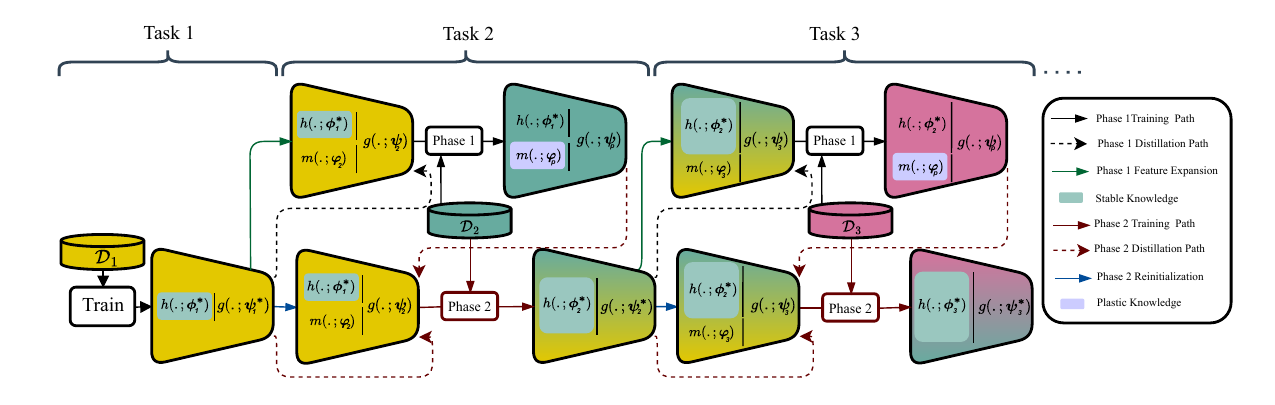}
\caption{ \textit{Phase~1} concludes by keeping the expanded module of $m(\cdot;\nhpr{t})$ as plastic knowledge for task $\mathcal{T}_t$.}
\label{fig:pkb-dyn}
\end{figure}

Considering expansion stage~\eqref{eq:dyn-expansion} in architecture-based methods, new module $m(\cdot;\nhpr{t})$ attempts to acquire task~$\mathcal{T}_t$'s representation in the presence of stable feature extractor $h(\cdot;\hpr{t-1}^*)$~\eqref{eq:skb-dyn}. While task~$\mathcal{T}_t$'s feature extractor $h(\cdot;\hpr{t})$ is updated on $\mathcal{D}_t$ during \textit{Phase~1}, its main part $h(\cdot;\hpr{t-1}^*)$ is frozen; therefore, the new module $m(\cdot;\nhpr{t})$ captures new representation. We define the plastic knowledge to be $m(\cdot;\nhpr{t})$, obtained at \textit{Phase~1}'s conclusion (see Figure~\ref{fig:pkb-dyn}): 
\begin{equation}
\label{eq:pkb-dyn}
     \mathcal{P}  \stackrel{\text{def}}{=}  m(\cdot;\nhpr{p}) = m(\cdot;\nhpr{t}) .
\end{equation}

\subsection{Flashback Learning Phases}
\label{sec:fl-steps}
With a clear definition of stable and plastic knowledge in \textsection~\ref{sec:skb} and \textsection~\ref{sec:pkb}, we explain how SKB and PKB are incorporated in FL phases to harmonize the stability-plasticity balance.
\subsubsection{Phase~1}
\label{sec:phase1}
This phase plays a preparation step for bidirectional regularization; we update the model here to rapidly learn $\mathcal{D}_t$ and then refine the plastic knowledge. We update the model by the same loss function as the host CL method for a limited number of epochs $E_1$. The loss function is split into two components;
\begin{equation}
\label{eq:l1}
    \mathcal{L}_1(\tta) =  \mathcal{L}_{c}(\tta) + \alpha_s \mathcal{L}_s(\tta)\;.
\end{equation}
The first component $\mathcal{L}_{c}(\tta)$ is the task objective function (\eg classification loss), which helps the model to learn task~$\mathcal{T}_t$. Stability loss $\mathcal{L}_s(\tta)$ takes stable knowledge $\mathcal{S}$ from SKB to regularize model updates in a unidirectional manner, and $\alpha_s$ is its scaler. Here, we explain how CL methods engage their available stable knowledge in the stability loss:

\begin{itemize}
\item \underline{Distillation methods}' stability loss utilizes stable knowledge~\eqref{eq:skb-distill} to preserve new model representations close to stable model at feature embeddings, logits, or even intermediate features. For instance, in LUCIR~\cite{hou2019learning}, the stability loss keeps the similarity between $f(\cdot;\mpr{})$'s and $f(\cdot;\mpr{s})$'s normalized outputs by cosine embedding loss. The general form (See \ref{sec:general-dist}) of stability distillation loss is:

\begingroup
\color{dp}
\begin{equation}
\label{eq:ls-distil}
    \mathcal{L}_s(\tta) =  \mathbb{E}_{\vx \sim \mathcal{X}_t} \Big[ \frac{1}{2} \big \lVert f(\vx;\tta) - f(\vx;\tta_s) \big \rVert^2 \Big]\;.
\end{equation}

\endgroup

\item \underline{Replay methods} consider a stability loss component in their objective function that takes the memory samples into joint training. For example, in DER~\cite{buzzega2020dark}, $\mathcal{L}_s(\tta)$ is the mean squared error (MSE) loss between the model's response $f(\vx; \tta)$ to samples in memory (\ie $\vx \in \mathcal{M}_{t-1}$) and the corresponding old logits $\lgt{s}$ in SKB~\eqref{eq:skb-replay}. So, the stability loss forces the model response to memory samples to remain the same as old responses,  
\begin{equation}
\label{eq:ls-replay}
    \mathcal{L}_s(\tta) = \mathbb{E}_{(\vx,\lgt{s}) \sim \mathcal{S}} \Big[ \frac{1}{2} \big \lVert \mdl{\vx}{}{} - \lgt{s} \big \rVert^2 \Big].
\end{equation}

\item \underline{Parameter regularization methods} put their stable knowledge~\eqref{eq:skb-reg} into a stability loss component to impose constraints on model updates so old critical parameters are preserved. The stability loss is the distance between the parameters of the model under training and the stable model weighted by the stable FIM as:
\begin{equation}
\label{eq:ls-reg}
     \mathcal{L}_s(\tta) = \frac{1}{2} \left(\tta - {\tta_s}\right)^\top \mathbf{F}_s \left(\tta - {\tta_s}\right). 
\end{equation}

\item \underline{Dynamic architecture methods} update task~$\mathcal{T}_t$'s expanded feature extractor~\eqref{eq:dyn-expansion}, while its stable feature extractor $h(\cdot;\hpr{s})$~\eqref{eq:skb-dyn} is frozen. The stability loss formulation depends on how they devise the method to compatibly integrate representation learned by the new module $m(\cdot;\nhpr{t})$ into $h(\cdot;\hpr{s})$; various forms of distillation loss are commonly used to maintain the similarity between probability distribution modeled by current model ($\mdl{\vx}{}{}$ after expansion) and stable model $\mdl{\vx}{s}{}$. Let  $\hat{\vy}(\vx;\tta) = \mathrm{Softmax}\big(\mdl{\vx}{}{}\big)$ and $\hat{\vy}(\vx;\tta_s) = \mathrm{Softmax}\big(\mdl{\vx}{s}{}\big)$ be the output of current and stable models after \textit{Softmax} layer, respectively. FOSTER~\cite{wang2022foster} employs the Kullback-Leibler~(KL) divergence loss between $\hat{\vy}(\vx;\tta)$ and $\hat{\vy}(\vx;\tta_s)$ as:
\begin{equation}
\label{eq:ls-dyn}
     \mathcal{L}_s(\tta) = -\mathbb{E}_{\vx \sim \mathcal{X}_t} \Big[ \hat{\vy}(\vx;\tta_s)^\top \log \hat{\vy}(\vx;\tta)\Big].
\end{equation}

Assuming a linear classifier, $\mdl{\vx}{}{}=g (h(\vx;\hpr{});\gpr{})= \gpr{} h(\vx;\hpr{})$ and the expanded feature extractor~\eqref{eq:dyn-expansion}; we decompose  task~$\mathcal{T}_t$'s model  to $\mdl{\vx}{}{}= \gpr{} \begin{bmatrix} h(\vx; \hpr{s}) \\ m(\vx; \nhpr{}) \end{bmatrix}$; and its stable model to $\mdl{\vx}{s}{}= \gpr{s} h(\vx; \hpr{s})$ to detail trainable parameter set as $\mpr{}=\{\gpr{},\nhpr{}\}$. Stability loss~\eqref{eq:ls-dyn} attempts to reduce the distance between the predicted distribution parameterized by $\mdl{x}{}{}$ and $\mdl{x}{s}{}$.     
\end{itemize}
To recap, \textit{Phase~1} updates the model on $\mathcal{D}_t$ for limited epochs $E_1$ with objective function $\mathcal{L}_1(\tta)$ \eqref{eq:l1}. We define this objective function per CL category in \eqref{eq:ls-distil}-\eqref{eq:ls-dyn}. We obtain primary model $\mdl{\cdot}{p}{}$ at the end of this phase, from which we derive plastic knowledge $\mathcal{P}$~\eqref{eq:pkb-distill}-\eqref{eq:pkb-dyn} for the next phase.

\subsubsection{Phase~2}
\label{sec:phase2}
Having refined plastic knowledge $\mathcal{P}$ compatible with the stable knowledge  $\mathcal{S}$ kept by the host CL method, we access two required sources, SKB and PKB, for bidirectional knowledge transfer. In this phase, we re-initialize the model to stable model $f(\cdot;\tta_s)$; and proceed with training on task $\mathcal{T}_t$ for a sufficient number of epochs $E_2$ by below loss function;
\begin{equation}
\label{eq:l2}
    \mathcal{L}_2(\tta) = \mathcal{L}_{c}(\tta) + \alpha_s \mathcal{L}_s(\tta) + \alpha_p \mathcal{L}_p(\tta).
\end{equation}

The first two terms of $\mathcal{L}_2(\tta)$ in \eqref{eq:l2}, task-specific objective function $\mathcal{L}_{c}(\tta)$ and stability loss $\mathcal{L}_s(\tta)$, are analogous to $\mathcal{L}_1(\tta)$ in \eqref{eq:l1}. While stability loss benefits from stable knowledge in SKB to enhance model stability, the additional term $\mathcal{L}_p(\tta)$ is introduced to take plastic knowledge $\mathcal{P}$ from PKB and adjust the model's plasticity; $\alpha_p$ is its scaler. Therefore, stability-plasticity balance is harmonized by the counteract between stability and plasticity loss components in a bidirectional flow. Here, we explain how the FL mechanism takes plastic knowledge into the plasticity loss:
\begin{itemize}
\item \underline{Distillation methods} attempt to keep model representations close to the stable model~\eqref{eq:skb-distill} in the stability loss~\eqref{eq:ls-distil}; so, we formulate the plasticity loss to concurrently keep model representation close to the primary model $\mdl{\cdot}{p}{}$ kept in the PKB~\eqref{eq:pkb-distill}: 
\begin{equation}
\label{eq:lp-distil}
    \mathcal{L}_p(\tta) =  \mathbb{E}_{\vx \sim \mathcal{X}_t} \Big[ \frac{1}{2} \big \lVert f(\vx;\tta) - f(\vx;\tta_p) \big \rVert^2 \Big]\;.
\end{equation} 

\item \underline{Replay methods'} stability loss~\eqref{eq:ls-replay} reduce the distance between the model response to memory samples $\vx \in \mathcal{M}_{t-1}$ and their corresponding old logits $\lgt{s}$ in SKB~\eqref{eq:skb-replay}. We define an MSE plasticity loss: 
\begin{equation}
\label{eq:lp-replay}
    \mathcal{L}_p(\tta) = \mathbb{E}_{(\vx,\lgt{p}) \sim (\mathcal{M}_{t-1} , \mathcal{P}) } \Big[ \frac{1}{2} \big \lVert \mdl{\vx}{}{} - \lgt{p} \big \rVert^2 \Big].
\end{equation}
to reduce the distance between model response to memory samples and their primary logits $\lgt{p}$ kept in its PKB~\eqref{eq:pkb-replay}.

\item \underline{Parameter regularization methods}'s stability loss~\eqref{eq:ls-reg} attempts to preserve important parameters of previous tasks. We take task~$\mathcal{T}_t$'s primary parameters and their FIM from PKB~\eqref{eq:pkb-reg} and define plasticity loss to follow the same format:  
\begin{equation}
\label{eq:lp-reg}
\mathcal{L}_p(\tta) =  \frac{1}{2} \left(\tta - {\tta_p}\right)^\top \mathbf{F}_p \left(\tta - {\tta_p}\right), 
\end{equation}
with a focus on critical parameters to task~$\mathcal{T}_t$. 

\item In \underline{dynamic architecture methods}, having kept a replica of the new feature extractor module~\eqref{eq:pkb-dyn} in PKB, we reinitialize $m(\cdot;\nhpr{})$ when \textit{Phase~2} begins. Corresponding to the stability loss~\eqref{eq:ls-dyn} defined in the host CL method to align task~$\mathcal{T}_t$ and stable model outputs, we define a plasticity loss to reduce the distance between the probability distribution parameterized by $\mdl{x}{}{}$ and primary new model $\mdl{x}{p}{}$:

\begin{equation}
\label{eq:lp-dyn}
\mathcal{L}_p(\tta) = -\mathbb{E}_{\vx \sim \mathcal{X}_t} \Big[ \hat{\vy}(\vx;\tta_p)^\top \log \hat{\vy}(\vx;\tta)\Big];
\end{equation}
in which $\hat{\vy}(\vx;\tta_p) = \mathrm{Softmax}\big(\mdl{\vx}{p}{}\big)$ is the output after softmax layer in primary new model~$\mdl{\vx}{p}{}$. With linear classifier and expansion strategy, the primary model is decomposed to $\mdl{\vx}{p}{}=\gpr{p} \begin{bmatrix} h(\vx; \hpr{s}) \\ m(\vx; \nhpr{p}) \end{bmatrix}$; so it has absorbed task~$\mathcal{T}_t$ knowledge in its new module kept in PKB~\eqref{eq:pkb-dyn}.
\end{itemize}
To summarize FL's \textit{Phase~2}, we reinitialize the model's parameters to the stable model's and update it on $\mathcal{D}_t$ by optimizing $\mathcal{L}_2(\tta)$~\eqref{eq:l2}. Stability and plasticity losses harness existing and refined new knowledge from SKB and PKB to guide model updates concurrently (see Algorithm~\ref{alg:Flashback}). 
\subsubsection{Training and Memory Budget}  

We design FL to maintain a training budget comparable to that of the host CL method. Given that the host CL method trains for \(E_{\text{CL}}\) epochs, we structure FL training in two phases with the same number of epochs. In \textit{Phase 1}, the model is trained on the current task \(\mathcal{T}_t\) for a small number of epochs (\(E_1 \ll E_{\text{CL}}\)), allowing it to quickly adapt. \textit{Phase 2} then resumes training from the old checkpoint for \(E_2 = E_{\text{CL}} - E_1\) epochs. This ensures that the total effective training epochs for FL remain the same as the host CL method, i.e.,  
\[
E_{\text{FL}} = E_1 + E_2 = E_{\text{CL}}.
\]  
In terms of memory overhead, FL approximately doubles the memory requirement of the host CL method, requiring a total memory budget of:  
\[
\mathbf{M}_{\text{FL}} = \mathbf{M}_\text{PKB} + \mathbf{M}_\text{SKB} = 2\mathbf{M}_\text{CL}.
\]  
For the \textbf{SKB}, FL follows the memory strategy of the host CL method. If the host CL method allocates \(\mathbf{M}_\text{CL}\) for storing old model checkpoints, parameters, or a subset of past samples, FL uses the same amount for SKB, ensuring  
\[
\mathbf{M}_\text{SKB} = \mathbf{M}_\text{CL}.
\]  
For the \textbf{PKB}, FL extracts plastic knowledge from the new model in a manner consistent with how the host CL method preserves stable knowledge. If the host method requires \(\mathbf{M}_\text{CL}\) to store stable knowledge, an equal amount is allocated for PKB, leading to  

\[
\mathbf{M}_\text{PKB} = \mathbf{M}_\text{CL}.
\]  

Thus, while FL introduces additional memory overhead, it does so in a structured and predictable manner, ensuring that training remains efficient without exceeding the computational constraints of the host CL method.

\begin{algorithm}[tb]
\caption{FL Algorithm}
\label{alg:Flashback}
\textbf{Input}: Trained model $\mdl{\cdot}{t-1}{*}$ till task~$\mathcal{T}_{t-1}$, Input data $(\vx,\vy)\in \mathcal{D}_t$ from task~$\mathcal{T}_t$, \textit{Phase~1} and \textit{Phase~2} epochs $E_1$, $E_2$, Available $\mathcal{S}$ from \textit{SKB} \\
\textbf{Output}: Trained model $\mdl{\cdot}{t}{*}$ after task~$\mathcal{T}_t$, Updated $\mathcal{S}$ for next task
\begin{algorithmic}[1] 
\STATE Let $\mdl{\cdot}{s}{}=\mdl{\cdot}{t-1}{*}$.  \textit{\textcolor{blue}{\small \# Let stable model be trained model till task~$\mathcal{T}_{t-1}$ }}
\STATE Let $\mdl{\cdot}{t}{}=\mdl{\cdot}{s}{}$. \textit{\textcolor{blue}{\small \# \textbf{Phase~1}, Initialization }}
\FOR{epoch = 1, $\cdots$ , $E_{1}$} 
\STATE Retrieve available $\mathcal{S}$ for CL method \textit{\textcolor{blue}{\small \#  $\mathcal{S}$ per CL category~\eqref{eq:skb-distill}, \eqref{eq:skb-replay}, \eqref{eq:skb-reg}, \eqref{eq:skb-dyn}}}
\STATE Use $\mathcal{S}$ to compute $\mathcal{L}_s(\tta)$ \textit{\textcolor{blue}{\small \#  $\mathcal{L}_s(\tta)$ per CL category~\eqref{eq:ls-distil}, \eqref{eq:ls-replay}, \eqref{eq:ls-reg}, \eqref{eq:ls-dyn}}}
\STATE Descend gradient on $\mathcal{L}_1(\tta)$ \textit{\textcolor{blue}{\small \#  $\mathcal{L}_1(\tta)$~\eqref{eq:l1}}}
\STATE Update model $\mdl{\cdot}{t}{}$ on $(\vx,\vy) \in \mathcal{D}_{t}$.
\ENDFOR
\STATE Let $\mdl{\cdot}{p}{}=\mdl{\cdot}{t}{}$. \textit{\textcolor{blue}{\small \# Let primary model be updated model after \textit{Phase~1} }}
\STATE Extract $\mathcal{P}$ using $\mdl{\cdot}{p}{}$ \textit{\textcolor{blue}{\small \# $\mathcal{P}$ per CL category~\eqref{eq:pkb-distill}, \eqref{eq:pkb-replay}, \eqref{eq:pkb-reg}, \eqref{eq:pkb-dyn}}}
\STATE Let $\mdl{\cdot}{t}{}=\mdl{\cdot}{s}{}$.  \textit{\textcolor{blue}{\small \# \textbf{Phase~2}, Re-initialization }}
\FOR{epoch = 1, $\cdots$ , $E_{2}$}
\STATE Retrieve $\mathcal{S}$ and corresponding $\mathcal{P}$ for CL method 
\STATE Use $\mathcal{S}$ to compute $\mathcal{L}_s(\tta)$ \textit{\textcolor{blue}{\small \#  $\mathcal{L}_s(\tta)$ per CL category~\eqref{eq:ls-distil}, \eqref{eq:ls-replay}, \eqref{eq:ls-reg}, \eqref{eq:ls-dyn}}}
\STATE Use $\mathcal{P}$ to compute$\mathcal{L}_p(\tta)$ \textit{\textcolor{blue}{\small \#  $\mathcal{L}_p(\tta)$ per CL category~\eqref{eq:lp-distil}, \eqref{eq:lp-replay}, \eqref{eq:lp-reg}, \eqref{eq:lp-dyn}}}
\STATE Descend gradient on $\mathcal{L}_2(\tta)$ \textit{\textcolor{blue}{\small \#  $\mathcal{L}_2(\tta)$~\eqref{eq:l2}}}
\STATE Update model $\mdl{\cdot}{t}{}$ on $(\vx,\vy) \in \mathcal{D}_{t}$.
\ENDFOR
\STATE Extract $\mathcal{S}$ using $\mdl{\cdot}{t}{*}$ 
\STATE \textbf{return} $\mathcal{S}$,  $\mdl{\cdot}{t}{*}$
\end{algorithmic}
\end{algorithm}

\begingroup
\color{dp}
\section{Stability-Plasticity Analysis}
\label{sec:analysis}
This section demonstrates how FL can affect stability and plasticity in continual learning. Consider a model with a parameter vector $\tta \in \mathbb{R}^p$. The aim is to update the model for task $\mathcal{T}_t$. We distinguish between two cases. In the CL setting, the updated model is obtained by minimizing: 
\begin{align*}
    \mathcal{L}_{\text{CL}}(\mpr{}) = \mathcal{L}_c(\mpr{}) + \alpha_s \mathcal{L}_s(\mpr{})\;.
\end{align*}
In contrast, and as alluded to earlier, for FL, we perform: 
\begin{align*}
    \mathcal{L}_{\text{FL}}(\mpr{}) = \mathcal{L}_c(\mpr{}) + \alpha_s \mathcal{L}_s(\mpr{}) + \alpha_p \mathcal{L}_p(\mpr{})\;.
\end{align*}

We show that including $\alpha_p \mathcal{L}_p(\mpr{})$ will enhance the balance between stability and plasticity. Our theoretical development is based on the analysis of the gradient behavior in $\mathcal{L}_{\text{CL}}(\tta)$ and $\mathcal{L}_{\text{FL}}(\tta)$ :
\begin{align*}
    \nabla_{\mpr{}}\mathcal{L}_{\text{CL}}(\tta) &= \nabla_{\mpr{}}\mathcal{L}_{\text{c}}(\tta) + \alpha_s \nabla_{\mpr{}}\mathcal{L}_{\text{s}}(\tta)\;,\\
    \nabla_{\mpr{}}\mathcal{L}_{\text{FL}}(\tta) &= \nabla_{\mpr{}}\mathcal{L}_{\text{c}}(\tta) + \alpha_s \nabla_{\mpr{}}\mathcal{L}_{\text{s}}(\tta) 
    + \alpha_p \nabla_{\mpr{}}\mathcal{L}_{\text{p}}(\tta)\;.
\end{align*}
%

\subsection{Distillation Methods}
In distillation methods, the form of stability~\eqref{eq:ls-distil} and plasticity~\eqref{eq:lp-distil} loss are: 
\begin{align*}
\mathcal{L}_s(\tta) &=  \mathbb{E}_{\vx \sim \mathcal{X}_t} \Big[ \frac{1}{2} \big \lVert f(\vx;\tta) - f(\vx;\tta_s) \big \rVert^2 \Big]\;,\\
\mathcal{L}_p(\tta) &=  \mathbb{E}_{\vx \sim \mathcal{X}_t} \Big[ \frac{1}{2} \big \lVert f(\vx;\tta) - f(\vx;\tta_p) \big \rVert^2 \Big]\;;
\end{align*}
which leads to the following theorem.

\begin{theorem}[Gradient Decomposition in Distillation Methods] \label{thm:thm_distl_informal_cl_fl}
Let $\mdl{\cdot}{}{}$,  $\mdl{\cdot}{s}{}$, and  $\mdl{\cdot}{p}{}$ denote a model, its stable version from SKB~\eqref{eq:skb-distill}, and its primary new version from PKB~\eqref{eq:pkb-distill} respectively; then the gradient of distillation-based methods at a sample $(\vx,\vy) \sim \data{t}$ can be decomposed as: 
\begin{align}
     \nabla_{\mpr{}}\mathcal{L}_{\text{CL}}(\tta) &=  \underbrace{\nabla_{\mpr{}}\mathcal{L}_{c}(\tta)}_{\text{Task-specific Gradient}} +  \alpha_s \underbrace{ \nabla_{\tta} f(\vx;\tta)^\top \left( f(\vx;\tta) - f(\vx;\tta_s) \right)   }_{\text{Stability Gradient}}\;. \label{eqn:grad_cl_decomp_distill_inf} \\
    \nabla_{\tta} \mathcal{L}_{\text{FL}}(\tta) &=  \underbrace{\nabla_{\tta} \mathcal{L}_{c}(\tta)}_{\text{Task-specific Gradient}} + \left(\alpha_s+\alpha_p\right) \underbrace{\nabla_{\tta} f(\vx;\tta)^\top \left( f(\vx;\tta) - \frac{\alpha_s f(\vx;\tta_s) + \alpha_p f(\vx;\tta_p)}{\alpha_s+\alpha_p} \right)  }_{\text{Gradient Interpolation}}\;. 
    \label{eqn:grad_fl_decomp_distill_inf}
\end{align}
See \textsection\ref{subsec:p-distillation} for the proof. 
\end{theorem}

The interpolation term in gradient~\eqref{eqn:grad_fl_decomp_distill_inf} drives the output $f(\vx;\tta)$ toward an interpolation between the stable response $f(\vx;\tta_s)$ and the primary new response $f(\vx;\tta_p)$. The stable response encapsulates information from the old tasks $\{\mathcal{T}_i\}_{i=1}^{t-1}$, while the primary model contains information from task~$\mathcal{T}_t$. Therefore, the interpolation term guides the model output toward a balance between $f(\vx;\tta_s)$ and $f(\vx;\tta_p)$ and will enhance the stability-plasticity trade-off.

\subsection{Replay Methods}

We turn our attention to replay methods and show similar behavior to what we observed in distillation methods that can emerge with FL. With stability~\eqref{eq:ls-replay} and plasticity~\eqref{eq:lp-replay} loss in the form of:
\begin{align*}
    \mathcal{L}_s(\tta) =& \mathbb{E}_{(\vx,\lgt{s}) \sim \mathcal{S}} \Big[ \frac{1}{2} \big \lVert \mdl{\vx}{}{} - \lgt{s} \big \rVert^2 \Big] \\
    \mathcal{L}_p(\tta) =& \mathbb{E}_{(\vx,\lgt{p}) \sim (\mathcal{M}_{t-1}, \mathcal{P})} \Big[ \frac{1}{2} \big \lVert \mdl{\vx}{}{} - \lgt{p} \big \rVert^2 \Big]
\end{align*}
the following theorem formalizes the behavior. 

\begin{theorem}[Gradient Decomposition in Replay Methods] \label{thm:thm_replay_informal_cl_fl}
Let $(\vx , \vy)$ be a sample from memory~$\mathcal{M}_{t-1}$, and $\lgt{s}$ and $\lgt{p}$ be its corresponding stable logit kept in SKB~\eqref{eq:skb-replay} and primary new logits kept in PKB~\eqref{eq:pkb-replay}, respectively. 
The gradient of replay methods can be decomposed into two terms for CL and FL as follows: 
\begin{align}
     \nabla_{\tta} \mathcal{L}_{\text{CL}}(\tta) &=  \underbrace{\nabla_{\tta} \mathcal{L}_{c}(\tta)}_{\text{Task-specific Gradient}} +  \alpha_s \underbrace{\nabla_{\tta} f(\vx;\tta)^\top \left( f(\vx;\tta) - \lgt{s} \right) }_{\text{Stability Gradient}}\;. \label{eqn:grad_cl_decomp_replay_inf} \\
    \nabla_{\tta} \mathcal{L}_{\text{FL}}(\tta) &=  \underbrace{\nabla_{\tta} \mathcal{L}_{c}(\tta)}_{\text{Task-specific Gradient}} +  \left(\alpha_s+\alpha_p\right) \underbrace{\nabla_{\tta} f(\vx;\tta)^\top \left( f(\vx;\tta) - \frac{\alpha_s \lgt{s} + \alpha_p \lgt{p}}{\alpha_s+\alpha_p} \right)  }_{\text{Gradient Interpolation}}\;.
    \label{eqn:grad_fl_decomp_replay_inf}
\end{align}
See \textsection\ref{subsec:p-replay} for the proof. 
\end{theorem}

Once again, we note that in FL, an interpolation term is used during training. Unlike distillation methods, this gradient interpolation term in replay methods is determined using the logits of the SKB~\eqref{eq:skb-replay} and PKB~\eqref{eq:pkb-replay}. If the logits $\lgt{s}$ and $\lgt{p}$ are equal, the gradient term simplifies to that of the CL methods. If the stable model is sufficiently rich to generate the logits for the new task accurately, then a simple regularization is all we need in FL. However, if this is not the case, the gradient updates are governed by both SKB~\eqref{eq:skb-replay} and PKB~\eqref{eq:pkb-replay} logits. The contributions of SKB and PKB are adjusted by the hyperparameters $\alpha_s$ and $\alpha_p$, providing the designer with enough flexibility to update the model.

\subsection{Parameter Regularization Methods}
In this part, we will develop the behavior of FL for regularization methods, focusing on the role of the FIMs kept in SKB~\eqref{eq:skb-reg} and PKB~\eqref{eq:pkb-reg}. The following theorem will materialize this, considering stability~\eqref{eq:ls-reg} and plasticity~\eqref{eq:lp-reg} loss of form:
\begin{align*}
    \mathcal{L}_s(\tta)=&  \frac{1}{2} \left(\tta - {\tta_s}\right)^\top \mathbf{F}_s \left(\tta - {\tta_s}\right), \\
    \mathcal{L}_p(\tta)=& \frac{1}{2} \left(\tta - {\tta_p}\right)^\top \mathbf{F}_p \left(\tta - {\tta_p}\right). \
\end{align*}

\begin{theorem}[Gradient Decomposition in Parameter Regularization Methods ] \label{thm:thm_reg_informal_cl_fl}
Let $\mathbf{F}_s$ and $\mathbf{F}_p$ be the FIM kept in SKB~\eqref{eq:skb-reg} and PKB~\eqref{eq:pkb-reg}, respectively. Furthermore, let $\tta_s$ and $\tta_p$ denote the parameters of the stable and primary new version of model $\mdl{\cdot}{}{}$, kept in SKB~\eqref{eq:skb-reg} and PKB~\eqref{eq:pkb-reg}, respectively.
Then the gradient can be decomposed into two terms for CL and FL as follows: 
\begin{align}
     \nabla_{\tta} \mathcal{L}_{\text{CL}}(\tta) &=  \underbrace{\nabla_{\tta} \mathcal{L}_{c}(\tta)}_{\text{Task-specific Gradient}} +  
       \underbrace{ \alpha_s \mathbf{F}_s \big(\tta - \tta_s  \big)}_{\text{Stability Gradient}}\;. \label{eqn:grad_cl_decomp_reg_inf} \\
    \nabla_{\tta} \mathcal{L}_{\text{FL}}(\tta) &=  \underbrace{\nabla_{\tta} \mathcal{L}_{c}(\tta)}_{\text{Task-specific Gradient}} +  
    \underbrace{ \alpha_s \mathbf{F}_s\big(\tta - \tta_s  \big) +   \alpha_p \mathbf{F}_p\big(\tta - \tta_p\big)  }_{\text{Gradient Interpolation}}\;.
    \label{eqn:grad_fl_decomp_reg_inf}
\end{align}
See \textsection\ref{subsec:p-regularize} for the proof. 
\end{theorem}

This implies that the parameter update for CL and FL has the following recursive forms (See \textsection\ref{subsec:update-regularize-p} for details):
\begin{align}
    \mathrm{CL~Setting: ~ } \tta^{(k)} & = \underbrace{\big(\mathrm{I} - \Gamma_s \big)^k \tta^{(0)}}_{\text{Initial Term}} - \eta \underbrace{\sum_{j=0}^{k-1} \big(\mathrm{I} - \Gamma_s \big)^{k-1-j}  \Big( \nabla_{\tta} \mathcal{L}_{c}(\tta)\bigg|_{\tta=\tta^{(j)}} \Big)}_{\text{Task-specific Term}} + \underbrace{\sum_{j=0}^{k-1} \big(\mathrm{I} - \Gamma_s \big)^{k-1-j} \big(\Gamma_s \tta_s \big)^{j}}_{\text{Stability Term}}, \label{eqn:param_mov_reg_cl} \\
    \mathrm{FL~Setting: ~ } \tta^{(k)} &= \underbrace{\big(\mathrm{I} - \Gamma_s - \Gamma_p \big)^k \tta^{(0)}}_{\text{Initial Term}} - \eta \underbrace{\Big( \sum_{j=0}^{k-1} \big( \mathrm{I} - \Gamma_s - \Gamma_p \big)^{k-1-j} \nabla_{\tta} \mathcal{L}_{c}(\tta)\bigg|_{\tta=\tta^{(j)}} \Big)}_{\text{Task-specific Term}} \notag \\ &+   \underbrace{\sum_{j=0}^{k-1} \big( \mathrm{I}- \Gamma_s - \Gamma_p \big)^{k-1-j} \big( \Gamma_s \tta_s + \Gamma_p \tta_p \big)^ {j}}_{\text{Interpolation Term}}. \label{eqn:param_mov_reg_fl}
\end{align}
where $\Gamma_s =  \eta \alpha_s \mathbf{F}_s $ and  $\Gamma_p = \eta \alpha_p \mathbf{F}_p $. The initial term in Eq.~\eqref{eqn:param_mov_reg_cl} and Eq.~\eqref{eqn:param_mov_reg_fl} controls how much of the initial condition~$\tta^{(0)}$ remains over time. When eigenvalues of $\mathrm{I}-\Gamma_s$ (or $\mathrm{I}-\Gamma_s - \Gamma_p$) are between 0 and 1, the initial term decays to zeros as $k\to \infty$. The task-specific term stabilizes learning by a sum over past gradients of $\mathcal{L}_c (\tta)$. The influence of past gradients also diminishes over time due to multiplication with $\big(\mathrm{I}-\Gamma_s\big)^{k-1-j}$ (or $\big(\mathrm{I} - \Gamma_s - \Gamma_p\big)^{k-1-j}$). The third term which includes $\big(\Gamma_s \tta_s \big)$ (or $\big(\Gamma_s \tta_s + \Gamma_p \tta_p \big)$) determines the steady state solution; and if $\big(\mathrm{I}-\Gamma_s\big)^{k-1-j}$ (or $\big(\mathrm{I} - \Gamma_s - \Gamma_p\big)^{k-1-j}$) decays over time, the final steady-state solution depends mainly on $\big(\Gamma_s \tta_s \big)$ in CL setting while on $\big(\Gamma_s \tta_s + \Gamma_p \tta_p \big)$ in FL setting. Thus, the SGD trajectory in the CL setting~\eqref{eqn:param_mov_reg_cl} shows how the parameter updates remain centered around $\tta_s$. Compared to the CL setting, the FL update rule~\eqref{eqn:param_mov_reg_fl} incorporates the interpolation term, balancing the influence of both $\tta_s$ and $\tta_p$ during learning.
%


\subsection{Dynamic Architecture Methods}

Recall the model expands its feature extractor~\eqref{eq:dyn-expansion} to learn the new task in dynamic architecture methods. The new module is updated, while the probability distribution parameterized by the expanded model is controlled by KL divergence loss against the model with stable feature extractor~\eqref{eq:skb-dyn}. In the FL case, the probability distribution is concurrently controlled by KL divergence loss against the model with primary new module~\eqref{eq:pkb-dyn}. Using the stability~\eqref{eq:ls-dyn} and plasticity~\eqref{eq:ls-dyn} loss in the form of KL loss:
\begin{align*}
    \mathcal{L}_s(\tta) =& -\mathbb{E}_{\vx \sim \mathcal{X}_t} \Big[ \hat{\vy}(\vx;\tta_s)^\top \log \hat{\vy}(\vx;\tta)\Big],\\
    \mathcal{L}_p(\tta) =& -\mathbb{E}_{\vx \sim \mathcal{X}_t} \Big[ \hat{\vy}(\vx;\tta_p)^\top \log \hat{\vy}(\vx;\tta)\Big];
\end{align*}
the following theorem can explain the gradient of CL and FL settings for dynamic architecture methods.

\begin{theorem}[Gradient Decomposition in Dynamic Architecture Methods ] \label{thm:thm_dynamic_informal_cl_fl}
Let $\hat{\vy}(\vx;\tta)$, $\hat{\vy}(\vx;\tta_s)$ and $\hat{\vy}(\vx;\tta_p)$ be the probability distributions parameterized by a model $\mdl{\cdot}{}{}$ after expansion~\eqref{eq:dyn-expansion}, its stable version $\mdl{\cdot}{s}{}$ with feature extractor~\eqref{fig:skb-dyn} and its primary version $\mdl{\cdot}{p}{}$ incorporating the primary new module~\eqref{eq:pkb-dyn}, respectively. The gradient for a sample $(\vx, \vy) \sim \mathcal{D}_t$ can be decomposed into two terms for CL and FL as follows: 
\begin{align}
     \nabla_{\mpr{}}\mathcal{L}_{\text{CL}}(\tta) &=  \underbrace{\nabla_{\mpr{}}\mathcal{L}_{c}(\tta)}_{\text{Task-specific Gradient}} + 
     \frac{\alpha_s}{\tau} \underbrace{\nabla_{\mpr{}} \mdl{\vx}{}{}^\top \big( \hat{\vy}(\vx;\tta) - \hat{\vy}(\vx;\tta_s) \big) }_{\text{Stability Gradient}}\;. \label{eqn:grad_cl_dynamic_inf} \\
    \nabla_{\mpr{}}\mathcal{L}_{\text{FL}}(\tta) &=  \underbrace{\nabla_{\mpr{}}\mathcal{L}_{c}(\tta)}_{\text{Task-specific Gradient}} + 
    \frac{\alpha_s + \alpha_p}{\tau} \underbrace{\nabla_{\mpr{}} \mdl{\vx}{}{}^\top \bigg( \hat{\vy}(\vx;\tta) - \frac{\alpha_s \, \hat{\vy}(\vx;\tta_s)+ \alpha_p \, \hat{\vy}(\vx;\tta_p)}{\alpha_s + \alpha_p} \bigg) }_{\text{Gradient Interpolation}} \;;
    \label{eqn:grad_fl_dynamic_inf}
\end{align}
where $\tau$ is the temperature parameter of the Softmax, see \textsection\ref{subsec:p-dynamic} for the proof. 
\end{theorem}
The interpolation term in gradient~\eqref{eqn:grad_fl_dynamic_inf} drives the probability distribution  $\hat{\vy}(\vx;\tta)$ toward an interpolation between probability distributions parameterized by the stable model $f(\cdot;\tta_s)$ with the stable feature extractor~\eqref{eq:skb-dyn} and the primary model  $f(\cdot;\tta_p)$ with the primary new module~\eqref{eq:pkb-dyn}. The stable output~$\hat{\vy}(\vx;\tta_s)$ represents old tasks distribution $\{\mathcal{T}_i\}_{i=1}^{t-1}$, while the primary output~$\hat{\vy}(\vx;\tta_p)$ encapsulates $\mathcal{T}_t$'s distribution. Therefore, the interpolation term guides $\hat{\vy}(\vx;\tta)$ toward a balance between probability distributions modeled by stable model $f(\vx;\tta_s)$ and primary new model $f(\vx;\tta_p)$ and will enhance the stability-plasticity trade-off.

\endgroup

\section{Related Work}
\label{sec:literature}

We divide CL methods into four categories~\cite{mundt2023wholistic} and discuss their \textit{recent advancements} here. Moreover, we explain \textit{how} \textbf{FL} \textit{mechanism enhances the balance between stability and plasticity} in each CL category. In the initial phase, the FL mechanism compiles new task information to a knowledge base, promoting plasticity. FL aligns the content of the plastic knowledge base with retained information from previous tasks in the CL methods. This alignment is pivotal, as the subsequent phase entails competing for stability and plasticity using equivalent information from each side. We outline how FL shapes the plastic knowledge base and implements bidirectional regularization across four categories of CL methods.

\subsection{Distillation}

 Distillation-based CL methods \cite{li2017learning, hou2019learning,douillard2020podnet,jimagingKaushik, roy2023l3dmc} typically preserve a copy of the old model as their stable knowledge. Distillation occurs at different model representations; for instance, \cite{li2017learning} proposed minimizing the mean square error at logits between the model under training and the stable model. \cite{simon2021learning, roy2023subspace} perform knowledge distillation on final feature maps before the classification layer. PODNet \cite{douillard2020podnet} extended knowledge distillation to intermediate feature maps, and AFC \cite{kang2022class} incorporated gradient information to weigh the feature distillation loss and preserve essential features for old tasks. Some approaches limit their access only to the new task's dataset \cite{zhu2021prototype}, while others  \cite{rebuffi2017icarl,knights2022incloud} combine distillation with memory replay, focusing only on a selected subset of old samples, or they use some synthesized old samples \cite{yin2020dreaming} for distillation. Some methods approach distillation without keeping the stable model by utilizing encoded old knowledge into gradients~\cite{li2023memory}, or gradient norm through a warm-up step for a balanced gradient distillation \cite{fini2020online}; however, the final aim is the same, ensuring stability by transferring knowledge from the previous tasks to the new task.

 A successful distillation strategy should ideally balance stability and plasticity in the model, although plasticity has yet to receive as sufficient attention as stability in the literature \cite{kim2023stability}. Recent advancements in distillation methods attempt to achieve a better stability-plasticity balance by new features' projection onto the old ones within several approaches. \cite{fini2022self} uses a predictor network to project the representations from the new feature space (new model's feature embeddings) to the old one to ensure this new feature space is at least as powerful as the old one by self-supervised learning. \cite{gomez2022continually} also allows more plasticity to learn new features by projected functional regularization. Although these projection ideas provide more plasticity for the new model, they cannot explicitly take control over plasticity. Even with learnable linear transformation \cite{gu2023preserving} to simultaneously allow the separability of old classes and provide enough space for a new direction, we don't have explicit control over plasticity. After adding \textbf{FL} to distillation methods, we train the model on new data and conclude \textit{Phase 1} by integrating this model as plastic knowledge. In \textit{Phase 2}, the model is reinitialized to its old weights and trained on new task data again. During this phase, knowledge is distilled from two sources into the model: one side distills from the stable model and the other transfers from the new model stored in the plastic knowledge base. Two-way distillation avoids the overemphasis on stability and balances the focus on stability and plasticity equally. 

\subsection{Memory Replay}
Replay methods \cite{rebuffi2017icarl, buzzega2020dark, boschini2022class} guide model updates on the new task using a kept memory from the previous tasks. \cite{buzzega2020dark, boschini2022class, riemer2018learning, caccia2021new} maintain limited selected samples of the earlier tasks with their corresponding logits or feature embeddings in memory. Various memory management strategies are presented in the literature to choose old samples for memory. Some methods like Rainbow memory~\cite{bang2021rainbow} use classification uncertainty to ensure diversity of selected samples, while others like \cite{liu2020mnemonics} and \cite{ho2023prototype} pick more representative samples by using prototypes from previously seen classes. Moreover, some \cite{aljundi2019task} prioritize complex samples with high errors in memory selection. \textcolor{dp}{Some replay methods enhance their sample selection strategy by incorporating current task information. MIR~\cite{aljundi2019online} selects and replays memory samples whose predictions are most negatively impacted by the current parameter updates. This is achieved by computing the gradient of the loss with respect to the model parameters before and after the update. However, the focus on revisiting knowledge at the highest risk of being forgotten may come at the expense of not learning new knowledge.}

Recent methods prefer keeping processed information in memory for replay; this information might include features \cite{caccia2020online}, \cite{iscen2020memory}, or logits by previous models. The obvious advantage is reducing the storage cost; ALIFE~\cite{oh2022alife} refines and memorizes features of earlier tasks, while \cite{wang2022memory} uses features processed by data compression. Some methods \cite{liu2020generative}, \cite{ye2020learning}, and \cite{van2020brain} leverage generative models to eliminate storage costs; they train a generative model on previous tasks, then synthesize old samples at the beginning of the new task and use them for replay. Replay methods use different types of generative models for this purpose, from variational autoencoders \cite{binici2022robust}, \cite{van2021class} to \textcolor{dp}{generative adversarial networks \cite{cong2020gan}, \cite{li2024learning}.}

Despite the variety of memory content in replay methods, the principle remains the same: leveraging stable knowledge kept in memory in the training objective of task $t$ to retain what the model has learned from $\left\{\mathcal{D}_i\right\}_{i=1}^{t-1}$. DER/DER++~\cite{buzzega2020dark} and X-DER \cite{boschini2022class} utilize the memory samples in cross-entropy and distillation loss during task $t$ training. \cite{douillard2020podnet} and \cite{hou2019learning} employ a distillation loss at the feature level on memory samples. GEM~\cite{lopez2017gradient} and A-GEM~\cite{chaudhry2018efficient} impose constraints on task $t$ optimization by memory samples. As the model learns the new task, replay methods ensure that its responses to memory samples remain consistent with the old predictions. The loss component taking memory content into the objective function directly influences the stability but not plasticity. However, the new model offers more accurate predictions at new logits. In \textbf{FL}, \textit{Phase~1} learns the new task and integrates the new model's predictions for memory samples into the plastic knowledge base. In \textit{Phase~2}, the two-way regularization mechanism adjusts to align model responses to memory samples with the old predictions at old logits while being consistent with the new model's predictions at new logits. This bidirectional regularization on memory samples brings balance to the stability-plasticity competition. 
 
\subsection{Parameter Regularization}

Regularization methods aim to understand how modifications to model parameters affect task losses; they restrict updates on critical parameters to previous tasks to avoid forgetting. One notable approach is EWC \cite{kirkpatrick2017overcoming} introducing the Fisher information matrix to calculate the importance of model parameters. A distance-based penalty term is defined between new and old parameters by applying the calculated Fisher information matrix. When updating the model on task $t$, essential parameters in tasks $\left\{ i \right\}_{i=1}^{t-1}$ incur a more significant penalty, resulting in less modification. \cite{jung2020continual} proposes an adaptive approach to measure how crucial is each network node for carrying out the previous tasks. \cite{lee2020continual} extends the Kronecker-factored approximation of the Hessian matrix to weigh the penalty term between new and old parameters. SI \cite{zenke2017continual}, and MAS \cite{aljundi2018memory} utilize estimated path integrals and gradient magnitudes to regulate model parameter changes. RWalk~\cite{chaudhry2018riemannian} combines the Fisher information matrix and online path integral for approximating parameter importance. Online EWC \cite{schwarz2018progress} presents a solution for efficiently compiling the Fisher information matrix. \textcolor{dp}{ Meta-NC~\cite{ran2024learning} leverages the importance of previously learned parameters to compute and preserve the optimal inter-class margin. This computation relies on parameters learned during the meta-training phase. By retaining these parameters, the model ensures that the margin remains optimal, even as new classes are introduced.} 

Regularization methods rely on old parameters and their importance in encouraging stability and preserving the learned information from $\left\{\mathcal{D}_i\right\}_{i=1}^{t-1}$ within the model. However, their strategy limits the model's ability to learn task $t$ and does not promote plasticity. Some recent approaches \cite{pelosin2022towards} attempt to enhance learning capacity by employing richer architectures like vision transformers, but the objective function incorporates unidirectional regularization on attention layers for stability, not plasticity. 
\textcolor{dp}{Temporal ensembles~\cite{pmlr-v232-soutif-cormerais23a} regularize the model weights by computing the exponential moving average (EMA) of the weights during training. This technique smooths out updates, providing a more stable set of weights for evaluation. By mitigating abrupt changes in the current model weights, temporal ensembles ensure more stable performance over time, focusing on stability rather than plasticity. Recent methods, such as La-MAML~\cite{gupta2007maml}, introduce meta-learning into the regularization-based category and aim to improve the stability-plasticity balance through a look-ahead step in the optimization process. La-MAML~\cite{gupta2007maml} anticipates the impact of current updates on future learning and adjusts the learning rates of individual parameters accordingly. This approach leverages prior knowledge of the current task to optimize model performance across all tasks. While using information from the current task does help improve CL performance, fine-grained control is required to modulate the learning rate for each parameter. This makes the algorithm computationally intensive and challenging to scale to more complex sequential tasks.} Having learned task $t$ quickly in \textit{Phase 1}, \textbf{FL} provides regularization methods with the new model parameters and their importance. \textit{Phase 2} leverages old plus new parameters and their importance in regulating parameter updates. FL introduces a bidirectional regularization to control model updates, forcing the model parameters to stay at a proper distance from old and new parameters.  

\begin{figure}
\centering
\includegraphics[width=0.5\textwidth]{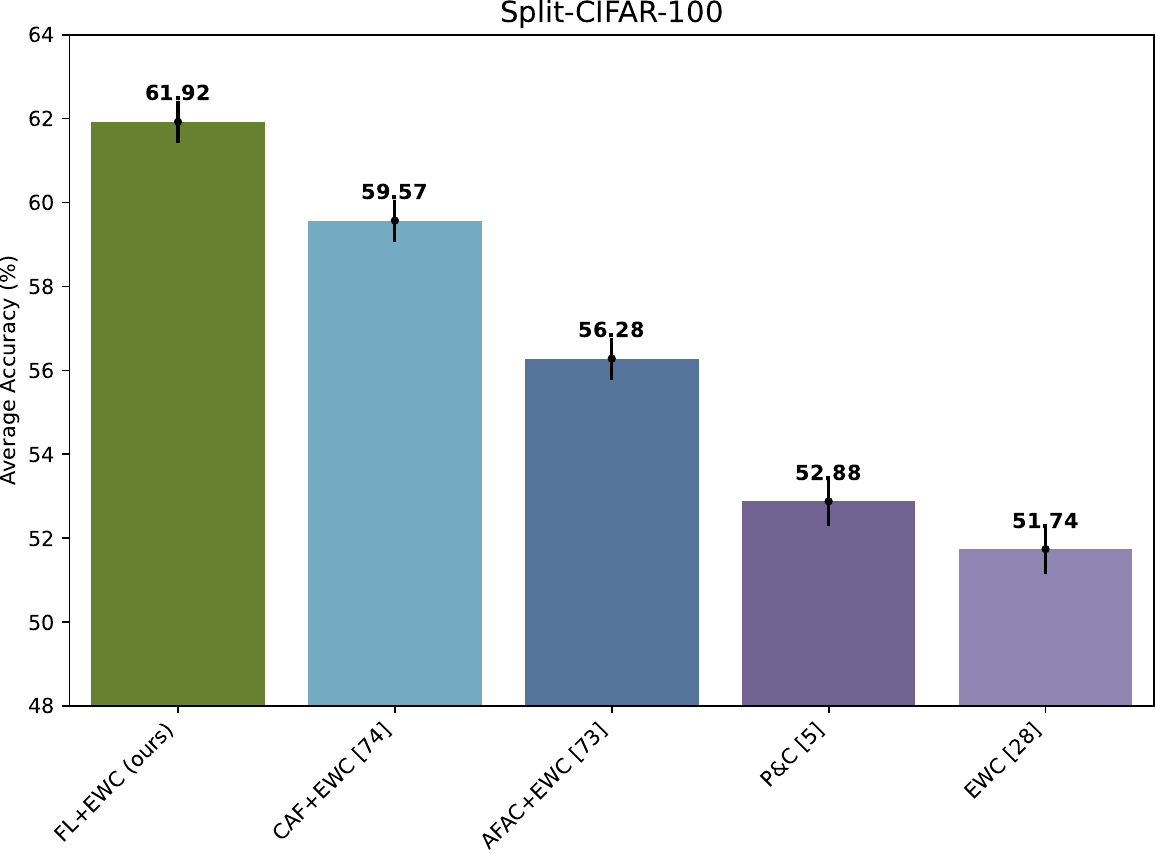}
\caption{\textbf{Average accuracy} (\% $\uparrow$) reported on Split CIFAR-100 benchmark for host CL baseline EWC~\cite{kirkpatrick2017overcoming} (from parameter regularization category) and its improvement by P\&C\cite{schwarz2018progress}, AFAC~\cite{wang2021afec}, CAF~\cite{wang2023incorporating}, and FL mechanism.}
\label{fig:bidirectional-comparison}
\end{figure}

\subsection{Dynamic Architecture Method}
Architecture-based methods assign a portion of the backbone to each task. This allocation typically occurs either by expanding the architecture at the start of a task $t$  \cite{yan2021dynamically}, \cite{wang2022foster}  or learning a suitable mask \cite{serra2018overcoming}, \cite{ostapenko2019learning} to activate a part of the model per task. The second approach often requires the task identifier $\left\{\ i \right\}_{i=1}^{\mathcal{T}}$ at the inference time to decide which mask to select, making this approach inapplicable to more challenging CL settings where a task identifier is unavailable. Some works extend the model plasticity by adding new neurons; \cite{yoon2018lifelong} evaluates loss function and decides to add a specific number of units to each layer when the task begins; or \cite{li2019learn} applies neural architecture search on a given topology, then optimizes the neuron connections to activate neurons gradually.

Some methods~\cite{yan2021dynamically} duplicate the entire backbone to expand the architecture, integrating a new learnable feature extractor into the old backbone and using channel-wise masking for pruning. Similarly, FOSTER~ \cite{wang2022foster} expands the backbone and then applies distillation to prevent redundancy in feature extraction and unused dimensions. BEEF~\cite{wang2023beef} outperforms FOSTER by independently training expanded modules and fusing them for bidirectional compatibility using an energy-based objective function. While dynamic expansion increases plasticity by providing additional capacity, model plasticity is not encouraged in the objective function. They usually freeze the old backbone and update the expanded module on task $t$ by knowledge transfer from old representations to maintain performance on tasks $\left\{ i \right\}_{i=1}^{t-1}$; however, this approach may hinder learning task $t$ effectively. In contrast, \textbf{FL} allows the extended module to learn task $t$ without constraints in \textit{Phase 1} and keeps the updated module. In \textit{Phase 2}, FL distills from the updated module into the model to counteract regularization by old representations and encourage plasticity in the objective function. 

\subsection{Bidirectional Knowledge Transfer for Stabililty-Plasticity Balance}
Conceptually close to our approach, P\&C~\cite{schwarz2018progress}, AFEC~\cite{wang2021afec}, CAF~\cite{wang2023incorporating}, and~\cite{lin2022towards} make use of bidirectional knowledge transfer to mitigate the stability-plasticity dilemma. However, they differ in how effectively they balance the two opposing forces. P\&C~\cite{schwarz2018progress} alternates between two distinct phases: Progress and Compress. In the Progress phase, only the active module is trained on the new task while linearly connected to the knowledge base. In the Compress phase, the knowledge base is updated via distillation from the active module and regularization against its previous state. While P\&C enables bidirectional knowledge flow, the interaction between the two modules is asymmetric, making it unclear how effectively each side influences the other.

AFEC~\cite{wang2021afec} addresses the rigidity of parameter preservation by introducing an active forgetting mechanism. Instead of strictly maintaining past knowledge, AFEC expands the model with new parameters, trains them independently on the new task, and then selectively merges them with existing parameters. Although designed as a plug-in method for CL, AFEC’s active forgetting mechanism operates independently of the host method, relying solely on parameter regularization rather than integrating with the host strategy’s knowledge transfer framework. CAF~\cite{wang2023incorporating} extends AFEC by employing a weighted average of multiple learners, each trained with a different forgetting rate. However, like AFEC, it primarily focuses on parameter regularization without explicitly incorporating the host CL strategy into its method.

Lin \etal \cite{lin2022towards}~takes a different approach by balancing stability and plasticity through two separately optimized networks. One network learns the current task via standard SGD, while the other preserves past knowledge through null-space projection. Unlike AFEC and CAF, this method is not a plug-in solution but a standalone framework, making it suitable for direct comparison against other CL baselines.

In contrast to these approaches, our FL algorithm is explicitly designed as a plug-and-play mechanism that integrates seamlessly with existing CL baselines. Unlike prior work, which either treats bidirectional knowledge flow asymmetrically or relies solely on parameter regularization, FL formulates knowledge transfer in alignment with the host method’s strategy. This ensures that both stability and plasticity counteract each other at the same level, achieving a better performance in the CL setting, which is evident in a head-to-head average accuracy comparison after augmentation with EWC~\cite{kirkpatrick2017overcoming} (Figure~\ref{fig:bidirectional-comparison}).


\section{Experiments}
\label{sec:experiment}

This section shows the results of adding FL to well-known CL baselines and recent state-of-the-art (SOTA) methods. In \textsection~\ref{sec:baseline}, we select at least one representative baseline from each CL category and add FL to observe its influence in more detail. Recent CL SOTA methods mainly benefit from a combination of CL strategies. In \textsection~\ref{sec:sota}, we integrate FL with those SOTA methods to assess whether it can improve their performance. \textcolor{dp}{Results from \textsection~\ref{sec:baseline} and \textsection~\ref{sec:sota} illustrate that the integration of the two-phase FL algorithm with existing CL methods promotes performance compared to when only CL methods are used, which might derive from the augmented strategy in FL. Hence, we also compared the performance of the FL mechanism to some recent plugin methods that adopt similar augmentation approaches on top of CL methods in \textsection~\ref{sec:plugin}. We study the effect of hyperparameters in the FL mechanism in \textsection~\ref{sec:hyperparam}. With additional experiments in~\textsection~\ref{sec:more-analysis}, we thoroughly analyze how the FL mechanism contributes to the stability-plasticity tradeoff.}

\subsection{Datasets}
\label{dataset}
We validate FL's performance on widely used datasets in CL for image classification tasks, including:
\begin{itemize}
    \item \textbf{CIFAR-10} \cite{krizhevsky2009learning} with ten classes, each class has 5000 and 1000 of $32\times32$ colored images, respectively, for training and testing.
    \item \textbf{CIFAR-100} \cite{krizhevsky2009learning} has 100 classes, and each class respectively comprises 500 and 100, train and test $32\times32$ colored images.
    \item \textbf{Tiny ImageNet} \cite{wu2017tiny} contains 100000 train and 10000 test downsized $64\times64$ colored images from ImageNet across 200 classes.
    \item \textbf{ImageNet-100} \cite{deng2009imagenet} is composed of 100 classes from the original ImageNet-1000 dataset.
\end{itemize} 

\subsection{Benchmarks}
\label{sec:bench}
For datasets listed in Section~\ref{dataset}, we consider the widely known protocols to create CL benchmarks 
\begin{itemize}
    \item For CIFAR-10, we divide ten classes into five tasks of two classes, and we train the model gradually with two classes per incremental task. This protocol is known as \textbf{Split CIFAR-10} in CL works \cite{rebuffi2017icarl},~\cite{zenke2017continual}. 
    \item CIFAR-100 appears in two different benchmarks:
    \begin{itemize}
        \item \textbf{CIFAR-100-B50-10}~(\cite{hou2019learning},\cite{simon2021learning},\cite{douillard2020podnet}): First task includes half of the classes (\ie fifty classes); and classes are divided into five ten-class tasks.
        \item \textbf{CIFAR-100-B0-10}~(\cite{rebuffi2017icarl}, \cite{boschini2022class}, \cite{wu2019large}): Classes are shown in ten tasks of ten classes. 
    
 \end{itemize}   
    \item In Tiny ImageNet, 200 classes are split into ten tasks of twenty classes. This benchmark is called \textbf{Split Tiny ImageNet} in \cite{buzzega2020dark},\cite{gu2023preserving}.
    \item ImageNet-100 are set into two benchmarks:
    \begin{itemize}
        \item \textbf{ImageNet100-B50-10}~(\cite{wang2022foster},\cite{simon2021learning},\cite{kang2022class}): The first task contains fifty classes, and the next five incremental tasks include ten classes each.
        \item \textbf{ImageNet100-B0-10}~(\cite{wang2023beef},\cite{wang2022foster}): 100 classes are shown in ten tasks of ten classes.
    \end{itemize}

\end{itemize}

\subsection{Metrics}
\begingroup
\color{dp}
To analyze the performance of 
 our method, we adopt common metrics used in the literature:
\begin{itemize}
    \item \textbf{Average Accuracy(AA)} shows the average accuracy across all tasks after sequential learning:
    \begin{equation}
        \label{eq:aa}
        AA = \frac{1}{T}\sum_{t=1}^{T} A_{T,t}
    \end{equation}
    where $A_{T,t}$ is the accuracy on task $t$ after learning all $T$ tasks.
    \item \textbf{Average Incremental Accuracy (AIA) } shows the average accuracy across all tasks after each task:
    \begin{equation}
        \label{eq:aia}
        \text{AIA} = \frac{1}{T}\sum_{t=1}^{T} \Big(\frac{1}{t}\sum_{j=1}^{t} A_{t,j} \Big)
    \end{equation}
    where $A_{t,j}$ is the accuracy on task $j$ after learning task $t$.
    \item \textbf{Forgetting (F)}~\cite{chaudhry2018riemannian} measures accuracy drop on previously learned tasks after learning $T$ tasks:
    \begin{equation}
        \label{eq:fg}
        F = \frac{1}{T-1} \sum_{t=1}^{T-1}\Big( \max_{k\in\{1,\dots,t\}} A_{k,t} - A_{T,t} \Big)
    \end{equation}
    where $\max_{k\in\{1,\dots,t\}} A_{k,t}$ is the best accuracy achieved on task $t$ at any prior learning step. 
\end{itemize}

We report the metrics mentioned above plus \textbf{Backward Transfer~(BWT)}~\cite{lopez2017gradient} and \textbf{Forward Transfer~(FWT)}~\cite{lopez2017gradient} to evaluate our experimental results. For a fair comparison with the CL methods, we used consistent metrics from their original papers: 
\begin{itemize}
    \item In \textsection~\ref{sec:baseline}, we use Average Accuracy~\eqref{eq:aa} to compare the model performance between \textbf{Method} and \textbf{Method+FL} ( Table~\ref{tab:CI}-\ref{tab:TI}.) This metric is a concise and immediate measure for comparisons among different baselines. In some papers \cite{wang2023beef}, it is denoted as Last Accuracy, although with the same definition as~\eqref{eq:aa}. In \textsection~\ref{sec:sota}, we add FL to \cite{wang2023beef} and \cite{wang2022foster} and assess its performance by this metric.  
    \item In \textsection~\ref{sec:plugin}, we compare FL with some recent two-phase plug-in CL methods \cite{kim2023achieving} and \cite{liu2021adaptive}. Consistent with the cited works, we take Average Incremental Accuracy~\eqref{eq:aia} to report and compare FL results (Table~\ref{tab:plugin}). 
    \item In \textsection~\ref{sec:ft-bt-results}, we evaluate Forgetting~\eqref{eq:fg}, Backward Transfer, and Forward Transfer before and after FL integration (Figure~\ref{fig:ft-bt}).
\end{itemize}
 
\endgroup

\begin{table}[ht]
  \centering
  \caption{\textbf{CI Setting}, FL integrated to five CL methods, selected methods are well-known representatives of three categories; \textbf{Distillation:} LUCIR and LwF.MC, \textbf{Parameter Regularization:} oEWC, \textbf{Memory Replay:} iCaRL and eXtended-DER. \textbf{Average Accuracy} ($\% \uparrow$) is reported for $\text{\textbf{Method}}$ and $\text{\textbf{Method}}\mathbf{+}\text{\textbf{FL}}$. (*) denotes results by the same implementation from \cite{buzzega2020dark} and \cite{boschini2022class}. \textcolor{teal}{\textbf{(Green)}} numbers in parentheses show improvement concerning baseline accuracy in each benchmark.}
  \resizebox{0.9\textwidth}{!}{
  \begin{tabular}{c|c|c|c}
    \hline
    \Xhline{1pt}
    \multirow{3}{*}{\textbf{Baselines}} & \multicolumn{3}{c}{\textbf{Class-incremental Setting}} \\
    \cline{2-4}
    & \textbf{\quad Split CIFAR-10 \quad} & \textbf{\quad Split CIFAR-100 \quad } & \textbf{\quad Split Tiny ImageNet \quad} \\
    & {\small 5 Tasks (2 Classes)} & {\small 10 Tasks (10 Classes)} & {\small 10 Tasks (20 Classes)} \\
    \hline
    \Xhline{1pt}
    \centering
    iCaRL~\cite{rebuffi2017icarl} & 70.21 & 49.82 * & 27.32 \\
    \cellcolor{mygray-bg} iCaRL+FL (ours) & \cellcolor{mygray-bg}  \textbf{72.39}~\textcolor{teal}{\textbf{(+2.18)}} & \cellcolor{mygray-bg}  \textbf{53.2}~\textcolor{teal}{\textbf{(+3.38)}} & \cellcolor{mygray-bg} \textbf{31.18}~\textcolor{teal}{\textbf{(+3.86)}} \\
    \hline
    LwF.MC~\cite{rebuffi2017icarl} & 43.14 & 16.22* & 24.18 \\
    \cellcolor{mygray-bg} LwF.MC+FL (ours) & \cellcolor{mygray-bg} \textbf{45.88}~\textcolor{teal}{\textbf{(+2.74)}} & \cellcolor{mygray-bg}  \textbf{21.13}~\textcolor{teal}{\textbf{(+4.91)}} & \cellcolor{mygray-bg}  \textbf{25.58}~\textcolor{teal}{\textbf{(+1.4)}} \\
    \hline
    oEWC~\cite{schwarz2018progress} &19.49* & 15.79 & 7.58* \\
    \cellcolor{mygray-bg} oEWC+FL (ours) & \cellcolor{mygray-bg} 
    \textbf{19.54}~\textcolor{teal}{\textbf{(+0.05)}} & \cellcolor{mygray-bg}  \textbf{16.34}~\textcolor{teal}{\textbf{(+0.55)}}  & \cellcolor{mygray-bg} \textbf{8.85}~\textcolor{teal}{\textbf{(+1.27)}}  \\
    \hline
     LUCIR~\cite{hou2019learning} & 79.89 &47.66 & 30.37 \\
     \cellcolor{mygray-bg}  LUCIR+FL (ours) & \cellcolor{mygray-bg} \textbf{80.91}~\textcolor{teal}{\textbf{(+1.02)}} & \cellcolor{mygray-bg}  \textbf{48.64}~\textcolor{teal}{\textbf{(+0.98)}} & \cellcolor{mygray-bg}  \textbf{32.47}~\textcolor{teal}{\textbf{(+2.10)}}\\
     \hline
     X-DER~\cite{boschini2022class} & 63.16 & 59.14* & 40.72\\
     \cellcolor{mygray-bg}  X-DER+FL (ours) & \cellcolor{mygray-bg}  \textbf{65.38}~\textcolor{teal}{\textbf{(+2.64)}} &  \cellcolor{mygray-bg} \textbf{58.02}~\textcolor{teal}{\textbf{(+0.45*)}} & \cellcolor{mygray-bg}  \textbf{41.81}~\textcolor{teal}{\textbf{(+1.09)}}\\
    \hline
    \Xhline{1pt}
  \end{tabular}
  }
  \label{tab:CI}
\end{table}
\subsection{Comparison with CL Baselines}
\label{sec:baseline}
This section lists some CL baselines to evaluate the FL effect. We chose at least one method from each CL category. These methods are well-known representatives of their category, allowing us to examine the impact of FL on different categories. Selected baselines are as follows;
\begin{itemize}
    \item \textbf{LUCIR}~\cite{hou2019learning}, \underline{L}earning a \underline{U}nified \underline{C}lassifier \underline{I}ncrementally via \underline{R}ebalancing, this baseline is more known for its distillation strategy that puts constraints on feature embeddings (before logits) to preserve knowledge. However, a replay of a set of selected exemplars is also allowed. 
    \item \textbf{iCaRL}~\cite{rebuffi2017icarl}, \underline{i}ncremental \underline{C}lassifier and \underline{R}epresentation \underline{L}earning, is a replay method with herding strategy to select samples closest to each class prototype and use them while distilling knowledge from the old model at logits. 
    \item \textbf{LwF}~\cite{li2017learning}, \underline{L}earning \underline{w}ithout \underline{F}orgetting, is a distillation baseline that tries to distill information from the old model into the new model at logits; in particular, we employ the \textbf{LwF.MC} an adaptation of LwF designed in \cite{rebuffi2017icarl}.
    \item \textbf{o-EWC}~\cite{schwarz2018progress}, \underline{o}nline \underline{E}lastic \underline{W}eight \underline{C}onsolidation is a baseline from parameter-regularization category, which calculates old parameters' importance recursively and then apply it for weighted regularization on new parameters update. 
    \item \textbf{X-DER}~\cite{boschini2022class}, e\underline{X}tended \underline{D}ark \underline{E}xperience \underline{R}eplay, is an extension to vanilla DER\cite{buzzega2020dark}, from replay category, that keeps old logits in the memory buffer for distillation during rehearsal. We selected X-DER because it performs better than other DER variations.   
\end{itemize}

\begin{table}[ht]
  \centering
  \caption{\textbf{TI Setting}, FL integrated to five CL methods, selected methods are well-known representatives of three categories; \textbf{Distillation:} LUCIR and LwF.MC, \textbf{Parameter Regularization:} oEWC, \textbf{Memory Replay:} iCaRL and eXtended-DER. \textbf{Average Accuracy} ($\% \uparrow$) is reported for $\text{\textbf{Method}}$ and $\text{\textbf{Method}}\mathbf{+}\text{\textbf{FL}}$. (*) denotes results by the same implementation from \cite{buzzega2020dark} and \cite{boschini2022class}. \textcolor{teal}{\textbf{(Green)}} numbers in parentheses show improvement concerning baseline accuracy in each benchmark.}
  \resizebox{0.9\textwidth}{!}{
  \begin{tabular}{c|c|c|c}
    \hline
    \Xhline{1pt}
    \multirow{3}{*}{\textbf{Baselines}} & \multicolumn{3}{c}{\textbf{Task-incremental Setting}} \\
    \cline{2-4}
    & \textbf{\quad Split CIFAR-10 \quad} & \textbf{\quad Split CIFAR-100 \quad } & \textbf{\quad Split Tiny ImageNet \quad} \\
    & {\small 5 Tasks (2 Classes)} & {\small 10 Tasks (10 Classes)} & {\small 10 Tasks (20 Classes)} \\
    \hline
    \Xhline{1pt}
    \centering
    iCaRL~\cite{rebuffi2017icarl} & 90.02 & 83.99  & 68.28 \\
    \cellcolor{mygray-bg} iCaRL+FL (ours) & \cellcolor{mygray-bg} \textbf{91.33 }~\textcolor{teal}{\textbf{(+1.31)}} & \cellcolor{mygray-bg} \textbf{86.77}~\textcolor{teal}{\textbf{(+2.78)}} & \cellcolor{mygray-bg}  \textbf{69.80}~\textcolor{teal}{\textbf{(+1.52)}} \\
    \hline
    LwF.MC~\cite{rebuffi2017icarl} &93.28 &61.40 & 65.45 \\
    \cellcolor{mygray-bg} LwF.MC+FL (ours) & \cellcolor{mygray-bg} \textbf{96.79}~\textcolor{teal}{\textbf{(+3.51)}} & \cellcolor{mygray-bg} \textbf{64.59}~\textcolor{teal}{\textbf{(+3.19)}} & \cellcolor{mygray-bg} \textbf{67.21}~\textcolor{teal}{\textbf{(+1.76)}} \\
    \hline
    oEWC~\cite{schwarz2018progress} & 68.29 & 59.34  & 19.20* \\
    \cellcolor{mygray-bg} oEWC+FL (ours) & \cellcolor{mygray-bg} \textbf{70.85}~\textcolor{teal}{\textbf{(+2.56)}} & \cellcolor{mygray-bg}  \textbf{62.35}~\textcolor{teal}{\textbf{(+3.01)}}  & \cellcolor{mygray-bg} \textbf{21.01}~\textcolor{teal}{\textbf{(+1.81)}}  \\
    \hline
     LUCIR~\cite{hou2019learning} & 96.61 & 86.88 & 69.20\\
     \cellcolor{mygray-bg} LUCIR+FL (ours) & \cellcolor{mygray-bg} 96.56~(-0.05)& \cellcolor{mygray-bg}\textbf{86.98}~\textcolor{teal}{\textbf{(+0.10)}}  &\cellcolor{mygray-bg} \textbf{69.69}~\textcolor{teal}{\textbf{(+0.49)}}  \\
     \hline
     X-DER~\cite{boschini2022class} & 92.86 & 88.89 & 75.16\\
     \cellcolor{mygray-bg}X-DER+FL (ours) & \cellcolor{mygray-bg}\textbf{95.51}~\textcolor{teal}{\textbf{(+2.65)}} & \cellcolor{mygray-bg}\textbf{89.49}~\textcolor{teal}{\textbf{(+0.60)}}  &\cellcolor{mygray-bg} \textbf{76.53}~\textcolor{teal}{\textbf{(+1.37)}}  \\
    \hline
    \Xhline{1pt}
  \end{tabular}
  }
  \label{tab:TI}
\end{table}

\subsubsection{Baseline Implementation}
It is important to note that comparing different baselines with intricate yet meaningful differences in experimental settings can be challenging. We built upon implementations of all baselines within a unified environment to address this issue, using available codebase \footnote{\url{https://github.com/aimagelab/mammoth}.} which allows comparison of different baselines on three commonly used benchmarks in CL setting, including \textbf{Split CIFAR-10}, \textbf{Split Tiny ImageNet}, and \textbf{Split CIFAR-100} ( this benchmark is same as \textbf{CIFAR-100-B0-10}). This approach ensures consistent evaluation of the FL mechanism across different methods. We use the ResNet18 \cite{he2016deep} for image classification following \cite{hou2019learning}, \cite{wu2019large}, \cite{castro2018end}, and \cite{buzzega2020dark}. The models are trained from scratch without any pre-training. Same as \cite{boschini2022class}, \cite{rebuffi2017icarl}, we employed a multi-epoch setup, allowing for multiple passes on the training data for each task. We use SGD with a predefined schedule for decreasing the learning rate at specific epochs. We keep the same value for the original CL and FL methods for main CL Hyperparameters like buffer size (in rehearsal methods), learning rate, or batch size. In LUCIR, XDER, and iCaRL, the buffer size for Split-CIFAR10 and Split-CIFAR100 is 2000 samples, and for Split Tiny ImageNet, it is 5120.  

\subsubsection{Baseline Results}

We present the Average Accuracy~\eqref{eq:aa} with/without FL for two main CL settings; see Table~\ref{tab:CI} for CI setting results and Table~\ref{tab:TI} for TI setting.  In these tables, accuracies reported with ${}^*$ are directly reported from \cite{buzzega2020dark} and \cite{boschini2022class}. In our experiments, we used the exact implementation as \cite{buzzega2020dark} and \cite{boschini2022class}, and when FL is integrated into CL methods, we keep the same hyperparameters for the CL part. Table~\ref{tab:CI} shows methods using memory replays, like iCaRL and X-DER, and distillation, like LwF.MC and LUCIR, for knowledge retention, are improved more than others by FL. The results show we have an improvement in accuracy in most benchmarks and models up to $\SI{+4.91}{\percent}$ in Split CIFAR-100 and up to $\SI{+3.86}{\percent}$ in Split Tiny ImageNet. Experimental results for X-DER show FL has improved average accuracy on all benchmarks except for Split-CIFAR100 (see Table~\ref{tab:CI} in the CI setting). We acknowledge that the source \cite{boschini2022class} reports an average accuracy of $\SI{59.15}{\percent}$, but we reproduced $\SI{57.57}{\percent}$ using the original code; then adding FL to the same setting, we improved accuracy from $\SI{57.57}{\percent}$ to $\SI{58.02}{\percent}$ which reflects an increase of $\SI{+0.45}{\percent}^*$.

In rehearsal-free methods like oEWC, there is a task bias toward recent classes in the absence of task identifier~\cite{ahn2021ss}, so original oEWC shows poor performance in the CI setting. As a common practice, we reported average accuracy after training on task $\mathcal{T}_T$, which shows up to $\SI{+1.27}{\percent}$ improvement for oEWC~(CI) on Tiny Imagenet; furthermore, throughout sequential training, adding FL boosts CL's average accuracy after each task. Considering average accuracies after each task $\{\mathcal{T}_t\}_{t=1}^T$ for oEWC and doing the same paired t-test analysis, we observed a p-value of $0.001$ and t-score $3.9$, showing CL+FL significance v.s CL.   

Table~\ref{tab:TI} shows improvements up to $\SI{3.19}{\percent}$, and $\SI{1.81}{\percent}$, respectively in Split CIFAR-100 and Split Tiny ImageNet. In the LUCIR method, the improvement margin is not as significant as that of other algorithms; LUCIR benefits from the dynamic expansion of the classifier head and fine-tuning of linear weights after representation learning, which helps it perform very well in the TI setting. So, the original LUCIR already obtained a high accuracy in the TI setting, and adding FL can just improve with a low margin. We did a paired t-test analysis to quantify the significance of FL on all methods in three benchmarks. See Table~\ref{tab:p-value} for positive t-scores and p-values~($<0.05$), showing CL+FL has statistically significant improvement over CL in evaluated methods, including all five algorithms included in Tables~\ref{tab:CI}-\ref{tab:TI}. 

\begin{table}[h]
    \centering
    \caption{T-test results, \textbf{p-value~$(\downarrow)$} and \textbf{t-score~$(\uparrow)$}, for \textbf{CL+FL} v.s. \textbf{CL} on three benchmarks for five selected methods from well-known CL baselines.}
    \label{tab:p-value}
    \resizebox{0.6\textwidth}{!}{
        \begin{tabular}{c|c|c|c}
        \toprule
        \textbf{Datasets} & \textbf{Split CIFAR-10} & \textbf{Split CIFAR-100} & \textbf{Split Tiny ImageNet} \\
        \midrule
        \textbf{p-value~$(\downarrow)$}~ &0.0009 & 0.001 & 0.0002  \\
        \midrule
        \textbf{t-score~$(\uparrow)$}  & 4.83 & 4.34 & 5.74 \\
        \bottomrule
        \end{tabular}
    }
\end{table}

\subsection{Comparison with SOTA Methods}
\label{sec:sota}
To show how FL can improve CL methods, we start the experiments with SOTA methods BEEF
~\cite{wang2023beef} and FOSTER~\cite{wang2022foster}, which benefit from a combination of different CL strategies, including distillation, memory replay, and architecture expansion to achieve the SOTA accuracy on standard CL benchmarks:
\begin{itemize}
    \item \textbf{BEEF}~\cite{wang2023beef}; \underline{B}i-compatible class-incremental learning via \underline{E}nergy-based \underline{E}xpansion and \underline{F}usion, it follows a two-stage training algorithm: expansion and fusion, the expansion stage is like the naive fine-tuning through energy-based optimization, in which they achieve bi-directional compatibility by adding backward (old classes confidence) and forward (open set uncertainty) prototypes and expanding $\mathcal{K}$ classification model into $\mathcal{K+2}$; the fusion stage is to use of old prototype confidence to take the final decision on all categories. 
    \item \textbf{FOSTER}~\cite{wang2022foster}; Feature Boosting and Compression for class-incremental learning, it is decoupled into two stages: boosting and compression; the boosting stage is to expand a trainable feature extractor and concatenate it to the frozen feature extractor of the old model (FOSTER uses distillation in boosting stage to instruct the expanded model, it also allows a fixed size memory buffer of old samples), the compression stage is for eliminating redundant parameters and dimensions caused by boosting.  
\end{itemize}

\begin{table}[ht]
  \centering
  \caption{Performance of FL on SOTA methods, \textbf{Average Accuracy} ($\% \uparrow$) for standard CL benchmarks is reported. The records for the methods in the first section of the table, and FOSTER, BEEF, and BEEF-Compress, are directly reported from \cite{wang2023beef}.}
  \resizebox{0.9\textwidth}{!}{
  \begin{tabular}{c|cc|cc}
    \hline
    \Xhline{1pt}
    \textbf{Method} & {\small\textbf{CIFAR-100-B0-10}} & {\small\textbf{CIFAR-100-B50-10}} & {\small\textbf{ImageNet-100-B0-10}} & {\small\textbf{ImageNet-100-B50-10}} \\
    \hline
    \Xhline{1pt}
    Bound & - & - & 81.50 & 81.50 \\
    \hline
    Replay\cite{riemer2018learning} & 41.01 & 41.26 & 41.00 & 43.38 \\
    iCaRL\cite{rebuffi2017icarl} & 49.52 & 52.04 & 50.98 & 53.69 \\
    BiC\cite{wu2019large} & 50.79 & 49.19 & 42.40 & 49.9 \\
    WA\cite{zhao2020maintaining} & 52.30 & 55.85 & 55.04 & 56.64 \\
    PodNet\cite{douillard2020podnet} & - & - & 45.40 & 62.94 \\
    DER\cite{yan2021dynamically} & 58.59 & 61.94 & 66.84 & 71.10 \\
    Dytox\cite{douillard2022dytox} & 57.76 & - & 57.94 & - \\
    \hline
    \Xhline{1pt}
    FOSTER\cite{wang2022foster} & 62.54 & 64.01 & 70.14 & 75.52 \\
    \cellcolor{mygray-bg} FOSTER+FL (ours) & \cellcolor{mygray-bg} \textbf{63.61} & \cellcolor{mygray-bg} 65.67 & \cellcolor{mygray-bg} 69.96 & \cellcolor{mygray-bg} 76.10 \\
    \hline
    \Xhline{1pt}
    BEEF\cite{wang2023beef} & 60.98 & 65.24 & 68.78 & 70.98 \\
    BEEF-Compress\cite{wang2023beef} & 61.45 & 64.54 & 71.12 & 74.62 \\
    \cellcolor{mygray-bg} BEEF-Compress+FL (ours) & \cellcolor{mygray-bg} 62.82 & \cellcolor{mygray-bg} \textbf{69.30} & \cellcolor{mygray-bg} \textbf{71.96} & \cellcolor{mygray-bg} \textbf{79.82} \\
    \hline
    \Xhline{1pt}
  \end{tabular}
  }
  \label{tab:sota}
\end{table}

\subsubsection{SOTA Implementation}
For adding FL to \textbf{BEEF}\cite{wang2023beef} and \textbf{FOSTER}\cite{wang2022foster}, we introduce \textit{Phase~1} of FL during BEEF's expansion stage (FOSTER's boosting stage, respectively) and train the expanded module on new data for a few epochs to obtain the primary version. In \textit{Phase~2} of FL, we incorporate distillation from the primary module to BEEF's forward-backward compatibility strategy (FOSTER's distillation strategy) to improve expansion and adapt to the new task. For implementation, we modified the available code base\footnote{\url{https://github.com/G-U-N/ICLR23-BEEF}.} to add the FL method. 
Following this implementation, we use ResNet18~\cite{he2016deep} for the backbone, SGD with the same learning and weight decay rates, Cosine scheduler, and a similar number of base epochs for a fair comparison. When a memory buffer is allowed, we use the same exemplar selection strategy and the same memory budget as the original CL method.

\begingroup
\subsubsection{SOTA Results}
\color{dp}
We report the quantitative results in Table~\ref{tab:sota} to show FL performance on SOTA methods on four main benchmarks \textbf{CIFAR-100-B0-10}, \textbf{CIFAR-100-B50-10}, \textbf{ImageNet-100-B0-10}, and \textbf{ImageNet-100-B50-10} defined in \textsection~\ref{sec:bench}. In this table, all experiments are reported in the CI setting. Results for CL methods in the first section of the table, FOSTER, BEEF, and BEEF-Compress are directly reported from \cite{wang2023beef} using the same implementation. The last Accuracy or Average Accuracy~\eqref{eq:aa} is reported in this table.
The results in Table~\ref{tab:sota} demonstrate that integrating FL with both FOSTER\cite{wang2022foster} and BEEF\cite{wang2023beef} consistently outperforms the SOTA methods across various benchmarks. \textit{FOSTER+FL} achieves higher accuracies on three out of four benchmarks, with notable improvements, including a 1.07\% increase on the first benchmark and a 1.66\% increase on the second. Although \textit{FOSTER+FL} did not exceed the reported SOTA accuracy on \textbf{ImageNet-100-B0-10} (reported in~\cite{wang2023beef} as 70.14\%), our replication using the FOSTER code\footnote{\url{https://github.com/G-U-N/ECCV22-FOSTER}.} yielded a baseline of 68.96\%, indicating that \textit{FOSTER+FL} still outperforms this result. 
Similarly, \textit{BEEF} shows significant gains after integration with FL, particularly on the second benchmark with a 4.76\% improvement and on the fourth benchmark with a 4.2\% increase. Even when augmented to BEEF-Compress\cite{wang2023beef}, which includes augmentation and memory management strategies, \textit{BEEF-Compress+FL} demonstrates superior performance, further validating the effectiveness of our approach.
\endgroup

\begingroup
\color{dp}
\subsection{Comparison with two-phase plugin methods}
\label{sec:plugin}
The findings from \textsection~\ref{sec:baseline} and \textsection~\ref{sec:sota} demonstrate that combining the two-phase FL algorithm with current CL methods enhances performance more than using CL methods alone. We further evaluate the FL mechanism's effectiveness against some recent plugin methods that implement comparable augmentation techniques alongside CL methods. We have chosen two plugin methods:
\setlength{\tabcolsep}{1.0mm}{
\begin{table*}[ht]
  \small
  \centering
    \caption{
  \textbf{Average incremental accuracies} (\% $\uparrow$) of iCaRL and LUCIR methods \emph{w/} and \emph{w/o} three plug-in methods: our \textbf{FL}, \textbf{ANCL}~\cite{kim2023achieving}, and \textbf{AANet}~\cite{liu2021adaptive}. Results for iCaRL and LUCIR (\emph{w/o} and \emph{w/} AANet) are directly reported from \cite{liu2021adaptive}. Results for iCaRL and LUCIR (\emph{w/} ANCL) are reported from \cite{kim2023achieving}.  
  {Please note \cite{kim2023achieving} didn't report the results for CIFAR-100-B50-2 benchmark }
}
  \vspace{-0.2cm}
  \begin{tabular}{lccccccccccc}
  \toprule
   \multirow{2.5}{*}{\textbf{\small{Method}}} & \multicolumn{3}{c}{\textbf{\small{Benchmark}} }  \\
  \cmidrule{2-4} 
   & \textbf{\footnotesize{CIFAR-100-B50-10}} & \textbf{\footnotesize{CIFAR-100-B50-5}}  & \textbf{\footnotesize{CIFAR-100-B50-2}}  \\
    \midrule

    iCaRL~\cite{rebuffi2017icarl} & 57.12\tiny{$\pm0.50$} & 52.66\tiny{$\pm0.89$} & 48.22\tiny{$\pm0.76$}  \\

    \cellcolor{mygray-bg}{\ \ \emph{w/} {ANCL}~\cite{kim2023achieving}} & \cellcolor{mygray-bg} 61.22\tiny{$\pm0.88$} & \cellcolor{mygray-bg}59.13\tiny{$\pm0.68$} &
    \cellcolor{mygray-bg} - \\
    {\ \ \emph{w/} {AANets}~\cite{liu2021adaptive} } & {64.22}\tiny{$\pm0.42$} &  {60.26}\tiny{$\pm0.73$} & {56.43}\tiny{$\pm0.81$}  \\
    \cellcolor{mygray-bg}{\ \ \emph{w/} {FL} } & \cellcolor{mygray-bg}{\highest{67.34}}\tiny{$\pm0.20$} &  \cellcolor{mygray-bg}{\highest{66.01}}\tiny{$\pm0.48$} &  \cellcolor{mygray-bg}{\highest{64.21}}\tiny{$\pm0.28$}  \\
    \midrule
    LUCIR~\cite{hou2019learning}  & 63.17\tiny{$\pm0.87$} & 60.14\tiny{$\pm0.73$} & 57.54\tiny{$\pm0.43$} \\
    
   \cellcolor{mygray-bg}{\ \ \emph{w/} {ANCL}~\cite{kim2023achieving}}  & \cellcolor{mygray-bg} 60.20\tiny{$\pm0.78$} & \cellcolor{mygray-bg} 60.04\tiny{$\pm0.8$} & \cellcolor{mygray-bg} - \\
   {\ \ \emph{w/} {AANets}~\cite{liu2021adaptive}}  & {66.74}\tiny{$\pm0.37$} & {65.29}\tiny{$\pm0.43$} & {63.50}\tiny{$\pm0.61$}  \\
    \cellcolor{mygray-bg}{\ \ \emph{w/} {FL}}  & \cellcolor{mygray-bg}{\highest{68.18}}\tiny{$\pm0.11$} & \cellcolor{mygray-bg}{\highest{67.23}}\tiny{$\pm0.13$} & \cellcolor{mygray-bg}{\highest{64.05}}\tiny{$\pm0.24$}  \\
  \bottomrule

\end{tabular}

\vspace{0.3cm}

  \label{tab:plugin}
  \vspace{-0.3cm}
\end{table*}
}
\begin{itemize}
    \item \textbf{AANet}~\cite{liu2021adaptive}; \underline{A}daptive \underline{A}ggregation \underline{Net}works for Class-Incremental Learning, it introduces two types of residual blocks—stable and plastic—at architecture during the first phase. Secondly, the outputs of these blocks are dynamically aggregated using adaptive weights to balance stability and plasticity effectively.
    \item \textbf{ANCL}~\cite{kim2023achieving}; \underline{A}uxiliary \underline{N}etwork \underline{C}ontinual \underline{L}earning, it introduces an auxiliary network in the first phase to enhance plasticity, then interpolates between stability and plasticity in the second phase.
\end{itemize}
In \cite{liu2021adaptive} and \cite{kim2023achieving}, respectively AANet and ANCL are augmented to standard CL baselines; they report their implementation results on several CL benchmarks, including:
\begin{itemize}
    \item CIFAR-100-B50-10 with \textbf{six} tasks: the first task has 50 classes, and the next five tasks contain ten classes each.
    \item CIFAR-100-B50-5 with \textbf{eleven} tasks: the first task has 50 classes, and the next ten tasks contain five classes each.
    \item CIFAR-100-B50-2 with \textbf{twenty-six} tasks: the first task has 50 classes, and the next twenty-five tasks contain two classes each.
\end{itemize}
We augment FL to their selected baselines with the same protocol and compare the performance in Table~\ref{tab:plugin}. The results are the average over three random seeds, and a similar strict memory budget is considered for fair comparison. In Table~\ref{tab:plugin}, iCaRL~\emph{w/}~FL shows an improvement of $7.78\%$ AIA~\eqref{eq:aia} compared to \cite{liu2021adaptive} in a challenging benchmark like CIFAR-100-B50-2  with a large number of tasks. In other benchmarks, iCaRL~\emph{w/}~FL and LUCIR~\emph{w/}~FL illustrate consistent improvement in AIA in comparison to results reported for ~\emph{w/}~AANet or ~\emph{w/}~ANCL. 

\endgroup
\subsection{Hyperparameters and Ablation Studies}
\label{sec:hyperparam}
In our experiments, hyperparameters include \textbf{CL Hyperparameters} and \textbf{FL Hyperparameters}. \textbf{CL Hyperparameters} are those shared with the original CL method, like learning rate, weight decay, buffer size, loss components scalers, and regularization scaling. \textbf{FL Hyperparameters} are those specific to the FL algorithm, including primary and final training epochs and plasticity loss scaler for bidirectional regularization. For CL baselines, we follow guidelines and recommendations reported in \cite{boschini2022class} to determine the value of CL Hyperparameters. When we integrate FL into each CL baseline, we do the same to specify CL Hyperparameters; then, we adjust the FL Hyperparameters empirically. For example, the original LUCIR has seven CL hyperparameters. We keep the CL hyperparameters the same as the baseline and focus on adjusting the FL hyperparameters, which contain three components: $E_1$, $E_2$ (opted to be same as host epochs in summation), and $\alpha_p$. \textcolor{dp}{In this section, we explore the effect of FL Hyperparameters on its performance.} 
\subsubsection{Ablatin study~-~FL epochs}
\begin{table}[h]
  \centering
  \caption{Ablation Study on $E_1$ and $E_2$: Average Accuracy~(\% $\uparrow$) for Split CIFAR-10 in CI setting.}
 
  \label{tab:hyper-E}
  \footnotesize
  \resizebox{0.5\columnwidth}{!}{
    \begin{tabular}{cccc} 
      \toprule
      \textbf{Scenario} & Hyperparameters & iCaRL\cite{rebuffi2017icarl} & LUCIR\cite{hou2019learning} \\
      &$E=E_1+E_2$ & &\\
      \midrule
      1 & Baseline~($E=60$)   & 70.21 & 79.89\\
      \midrule
      2 & Baseline+FL~($E=10+50$)   & \textbf{72.39} & \textbf{80.91} \\
      \midrule
      3 & Baseline+FL~($E=50+50$)   & 72.36 & 79.92\\
      \midrule
      4 & Baseline+FL~($E=10+10$)   & 69.88 & 72.59\\
      \midrule
      5 & Baseline+FL~($E=50+10$)   & 68.41 & 69.15\\
      \bottomrule
    \end{tabular}
  }
\end{table}

Ideally, we want FL not to demand more training epochs than the host algorithm (which FL integrated to); hence, we opted for $E_1+E_2$ to be similar to the host method's epochs. To illustrate the influence of these hyperparameters, we conducted a few experiments and compared the Average Accuracy against different combinations of $E=E_1+E_2$ in Table~\ref{tab:hyper-E}. The results confirm our intuition behind FL formulation; in \textit{Phase~1}, we need to learn new knowledge for a limited number of epochs, and in \textit{Phase~2}, we need to update the model for an adequate number of epochs by bidirectional regularization. In scenarios 4 and 5 in Table~\ref{tab:hyper-E} after absorbing new knowledge in \textit{Phase~1}, 10 epochs is not adequate to harmonize stability-plasticity balance in \textit{Phase~2}. While in scenarios 2 and 3 this balance is achieved in \textit{Phase~2}, excessively absorbing new knowledge during \textit{Phase~1} in scenario 3, might affect the balance negatively.
\begingroup
\color{dp}
\subsubsection{Ablatin study~-~Plasticity loss scaler}
\begin{table}[h]
  \centering
  \caption{Ablation Study on $\alpha_p$: Average Accuracy~(\% $\uparrow$) for Split CIFAR-10 in CI setting.}
  \label{tab:hyper-Ap}
  \footnotesize
  \resizebox{0.5\columnwidth}{!}{
    \begin{tabular}{cccc} 
      \toprule
      \textbf{Scenario} & Hyperparameters & iCaRL\cite{rebuffi2017icarl} & LUCIR\cite{hou2019learning} \\
      &$\alpha_p$ & &\\
      \midrule
      1 & Baseline~($\alpha_p=0$)   & 70.21 & 79.89\\
      \midrule
      2 & Baseline+FL~($\alpha_p=0.001$)   & \textbf{73.52} & 80.41 \\
      \midrule
      3 & Baseline+FL~($\alpha_p=0.01$)   & 72.39 & \textbf{80.91}\\
      \midrule
      4 & Baseline+FL~($\alpha_p=0.1$)   & 69.57 & 79.53\\
      \midrule
      5 & Baseline+FL~($\alpha_p=1$)   & 69.19 & 79.3 \\
      \bottomrule
    \end{tabular}
  }
\end{table}

Table \ref{tab:hyper-Ap} illustrates the impact of varying the hyperparameter $\alpha_p$ on the average accuracy of iCaRL and LUCIR methods for the Split CIFAR-10 dataset. Starting with $\alpha_p=0$, which represents the CL baseline with unidirectional regularization, we observe the initial accuracies of 70.21\% for iCaRL and 79.89\% for LUCIR. As we increase $\alpha_p$, we engage the FL bidirectional regularization, leading to an increase in average accuracy. This indicates that FL integration aids in better acquisition of new tasks and achieves a better balance with stability. The maximum average accuracy for iCaRL occurs at $\alpha_p=0.001$ with 73.52\%, while for LUCIR, it is at $\alpha_p=0.01$ with 80.91\%. However, further increasing $\alpha_p$ results in a drop in accuracy, which can be attributed to the model's tendency to focus more on learning new tasks at the expense of forgetting old tasks.

\subsubsection{Ablatin study~-~Memory Buffer}

\setlength{\tabcolsep}{1.0mm}{
\begin{table*}[ht]
  \small
  \centering
    \caption{
  \textbf{Average accuracy} (\% $\uparrow$) and \textbf{Forgetting} (\% $\downarrow$) of iCaRL \emph{w/} and \emph{w/o} FL reported on three benchmarks (\ie Split~CIFAR-10, Split~CIFAR-100, Split~Tiny~ImageNet) under four different memory budget: 200, 500, 2000, and 5120.   
}
  \vspace{-0.2cm}
  \begin{tabular}{lccccccccc}
  \toprule
   \multirow{2.5}{*}{\footnotesize{\textbf{Method}}} & \multirow{2.5}{*}{\footnotesize{\textbf{Memory Size}}} & \multicolumn{2}{c}{\footnotesize{\textbf{Split CIFAR-10}}} & \multicolumn{2}{c}{\footnotesize{\textbf{Split CIFAR-100}}} & \multicolumn{2}{c}{\footnotesize{\textbf{Split Tiny ImageNet}}} \\
  \cmidrule(lr){3-4} \cmidrule(lr){5-6} \cmidrule(lr){7-8}
   & & \footnotesize{\textbf{AA}~$\uparrow$} & \footnotesize{\textbf{F}~$\downarrow$} & \footnotesize{\textbf{AA}~$\uparrow$} & \footnotesize{\textbf{F}~$\downarrow$} & \footnotesize{\textbf{AA}~$\uparrow$} &  \footnotesize{\textbf{F}~$\downarrow$} \\
    \midrule
    \footnotesize
    iCarL~\cite{rebuffi2017icarl} & 200 & 65.91 & 22.46 & 39.94 & 34.50 & 19.17 & 33.07 \\
    \footnotesize
    \cellcolor{mygray-bg}{\ \ \emph{w/} FL} & \cellcolor{mygray-bg} 200 & \cellcolor{mygray-bg} 66.77 & \cellcolor{mygray-bg} 21.47 & \cellcolor{mygray-bg} 41.28 & \cellcolor{mygray-bg} 28.04 & \cellcolor{mygray-bg} 19.97 & \cellcolor{mygray-bg} 29.8 \\
    \midrule
    \footnotesize
    iCarL~\cite{rebuffi2017icarl} & 500 & 68.94 & 21.76 & 44.94 & 31.71 & 25.45 & 28.61 \\
    \footnotesize
    \cellcolor{mygray-bg}{\ \ \emph{w/} FL} & \cellcolor{mygray-bg} 500 & \cellcolor{mygray-bg}69.08 &  \cellcolor{mygray-bg} 20.12 & \cellcolor{mygray-bg} 48.19 & \cellcolor{mygray-bg} 24.51 & \cellcolor{mygray-bg} 26.42 & \cellcolor{mygray-bg} 26.26 \\
    \midrule
    \footnotesize
    iCarL~\cite{rebuffi2017icarl} & 2000 & 70.21 & 19.07 & 49.82 & 27.47 & 32.57 & 21.61 \\
    \footnotesize
    \cellcolor{mygray-bg}{\ \ \emph{w/} FL} & \cellcolor{mygray-bg} 2000 & \cellcolor{mygray-bg} 72.39 & \cellcolor{mygray-bg} 17.27 & \cellcolor{mygray-bg} 53.2&
    \cellcolor{mygray-bg} 23.07 &  \cellcolor{mygray-bg}33.83 & \cellcolor{mygray-bg} 19.52 \\
    \midrule
    \footnotesize
    iCarL~\cite{rebuffi2017icarl} & 5120 & 78.09 & 14.98 & 57.1 & 20.65 & 33.18 & 19.78 \\
    \footnotesize
    \cellcolor{mygray-bg}{\ \ \emph{w/} FL} & \cellcolor{mygray-bg} 5120 & 
    \cellcolor{mygray-bg} 78.20 & 
    \cellcolor{mygray-bg} 14.96 &
    \cellcolor{mygray-bg} 57.37 &
    \cellcolor{mygray-bg} 18.31 & 
    \cellcolor{mygray-bg} 35.16 &
    \cellcolor{mygray-bg} 18.68 \\
  \bottomrule
  \end{tabular}
  \label{tab:buffer}
\end{table*}
}

In CL baseline experiments~\textsection~\ref{sec:baseline}, we adhered to the typical buffer size settings from the original CL methods under the replay-based category (2000 for Split CIFAR-10 and Split CIFAR-100, and 5120 for Split Tiny ImageNet). We conducted additional experiments on these three benchmarks with varying memory budgets to assess FL effectiveness under limited memory conditions. Table~\ref{tab:buffer} presents the Average Accuracy~\eqref{eq:aa} and Forgetting~\eqref{eq:fg} in the CI setting for each benchmark under different memory sizes, both without and with FL integration. Our results indicate that the average accuracy is consistently higher, and the forgetting metric is lower in scenarios with FL. Thus, even under constrained memory conditions, incorporating FL into replay methods enhances average accuracy and reduces forgetting.

\subsubsection{Ablatin study~-~Architecture}

In our baseline experiments~\textsection~\ref{sec:baseline}, we used the standard ResNet18 backbone, consistent with prior works~\cite{hou2019learning}, \cite{wu2019large}, \cite{castro2018end}, and \cite{buzzega2020dark}. To further evaluate the robustness of our method, we conducted additional experiments using different backbone architectures. We employed the vision transformers~\cite{dosovitskiy2020image} with \textit{vit-small-patch16-224} and \textit{vit-tiny-patch16-224}. We integrated our method with iCaRL~\cite{rebuffi2017icarl} on the Split~CIFAR-100 benchmark. The results in Table~\ref{tab:arch} demonstrate that our method is agnostic to the architecture change, maintaining its effectiveness across different backbone models.
\setlength{\tabcolsep}{1.0mm}{
\begin{table*}[ht]
  \small
  \centering
    \caption{
  \textbf{Average accuracy} (\% $\uparrow$) and \textbf{Forgetting} (\% $\downarrow$) of iCaRL \emph{w/} and \emph{w/o} FL reported on  Split~CIFAR-100 with different backbones.  
}
  \vspace{-0.2cm}
  \begin{tabular}{lccccccccc}
  \toprule
   \multirow{2.5}{*}{\footnotesize{\textbf{Method}}}  & \multicolumn{2}{c}{\footnotesize{\textbf{ResNet-18}}} & \multicolumn{2}{c}{\footnotesize{\textbf{ViT-Small}}} & \multicolumn{2}{c}{\footnotesize{\textbf{ViT-Tiny}}} \\
  \cmidrule(lr){2-3} \cmidrule(lr){4-5} \cmidrule(lr){6-7}
   & \footnotesize{\textbf{AA}~$\uparrow$} & \footnotesize{\textbf{F}~$\downarrow$} & \footnotesize{\textbf{AA}~$\uparrow$} & \footnotesize{\textbf{F}~$\downarrow$} & \footnotesize{\textbf{AA}~$\uparrow$} &  \footnotesize{\textbf{F}~$\downarrow$} \\
    \midrule
    \footnotesize
    iCarL~\cite{rebuffi2017icarl} & 49.82 & 27.47 & 76.28 & 12.97 & 65.18 & 16.18  \\
    \footnotesize
    \cellcolor{mygray-bg}{\ \ \emph{w/} FL} & \cellcolor{mygray-bg} 53.2 & \cellcolor{mygray-bg} 23.07 & \cellcolor{mygray-bg} 79.21 & \cellcolor{mygray-bg} 8.6 & \cellcolor{mygray-bg} 66.22 & \cellcolor{mygray-bg} 12.74 \\
  \bottomrule
  \end{tabular}
  \label{tab:arch}
\end{table*}
}
\endgroup

\subsection{More Analysis}
\label{sec:more-analysis}
\subsubsection{Forgetting, Forward and Backward Transfer Analysis}
\label{sec:ft-bt-results}
\begin{figure}[ht]
\centering
\includegraphics[width=0.55\textwidth]{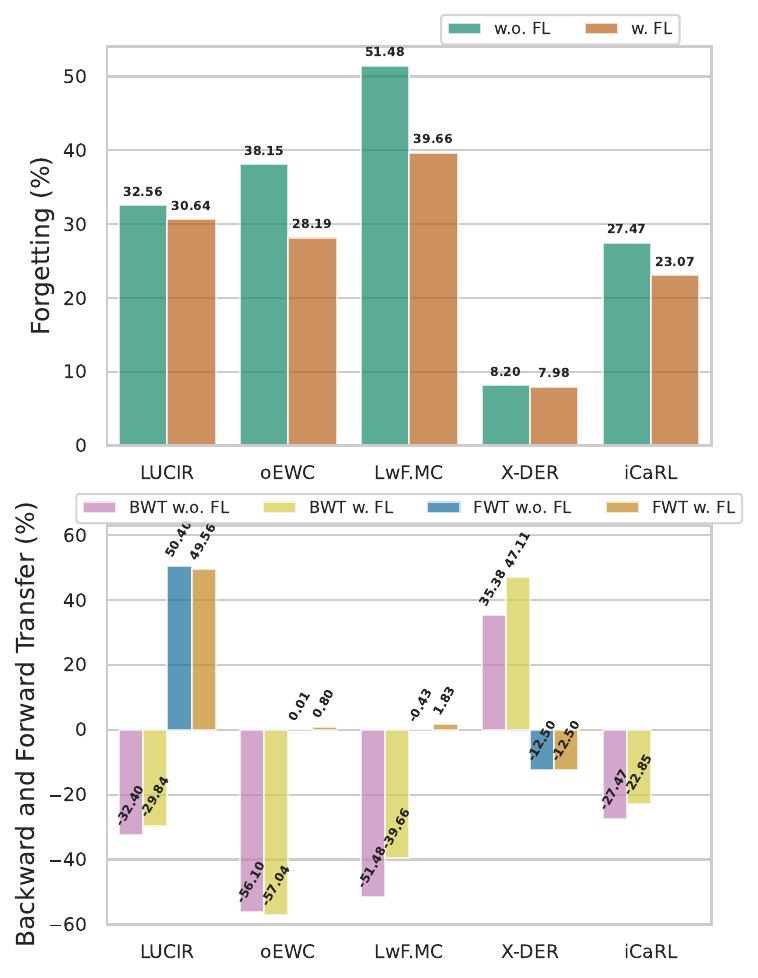} 
\caption{Average Forgetting ($\downarrow$), Backward Transfer ($\uparrow$) and Forward Transfer ($\uparrow$) are reported on Split-CIFAR-100.}
\label{fig:ft-bt}
\end{figure}

We analyze FL's impact on preserving knowledge by two main metrics: Forgetting and BWT. The first row of Figure~\ref{fig:ft-bt} reports Forgetting on Split CIFAR-100 benchmark under two scenarios: \textit{w.o.~FL} and \textit{w.~FL}. We observe that FL can reduce Forgetting, confirming that the bidirectional regularization has improved knowledge retention, while previously shown to improve Average Accuracy~(Table~\ref{tab:CI}-\ref{tab:TI}) as well. 

The Second row of Figure~\ref{fig:ft-bt} shows the BWT and FWT for five selected methods (FWT for iCaRL is not applicable for it uses Nearest Mean Exemplar classifier \cite{rebuffi2017icarl}). Large values of BWT and FWT show that the model has reinforced the accuracy of old and future classes in sequential training, respectively. Comparing BWT, we can see that, except for oEWC, other methods take larger BWT in augmentation with the FL mechanism. This observation confirms that learning task~$t$ after integration with FL has a less negative effect (or more positive effect) on old tasks' accuracy. Methods like oEWC and LwF.MC without memory samples takes a very small value FWT, although comparing these tiny values, the FL mechanism also positively influences future tasks' accuracy; we have the same interpretation for FWT in LUCIR, and no impact is shown on X-DER.

\begingroup
\color{dp}
\subsubsection{Stability-Plasticity Ratio}
\label{sec:spr-analysis}

\setlength{\tabcolsep}{1.0mm}{
\begin{table*}[ht]
  \small
  \centering
 \caption{
 \textbf{Stability-Plasticity Ratio}(\% $\downarrow$) of four CL representative methods \emph{w/} and \emph{w/o} our FL mechanism.}
  \vspace{-0.2cm}
  \resizebox{0.4\textwidth}{!}{
  \begin{tabular}{lccccccccccc}
  \toprule
   \multirow{2.5}{*}{\textbf{Method}} & \multicolumn{2}{c}{\textbf{Benchmark} }  \\
  \cmidrule{2-3} 
   & \textbf{\footnotesize{Split CIFAR-10}} & \textbf{\footnotesize{Split CIFAR-100}}    \\
    \midrule
    iCaRL~\cite{rebuffi2017icarl} & 21.87 &  37.47  \\
    \cellcolor{mygray-bg}{\ \ \emph{w/} {FL} } & \cellcolor{mygray-bg}{\highest{20.04}} & \cellcolor{mygray-bg}{\highest{33.07}} \\
    \midrule
   LwF.MC~\cite{rebuffi2017icarl} & 59.02 & 63.93  \\
    \cellcolor{mygray-bg}{\ \ \emph{w/} {FL} } & \cellcolor{mygray-bg}{\highest{55.36}} &  \cellcolor{mygray-bg}{\highest{61.21}} \\
    \midrule
    LUCIR~\cite{hou2019learning}  & 18.54 & 38.59\\    
   \cellcolor{mygray-bg}{\ \ \emph{w/} {FL}}  & \cellcolor{mygray-bg}{\highest{14.99}} & \cellcolor{mygray-bg}{\highest{36.54}} \\
    \midrule
   oEWC~\cite{schwarz2018progress}  & 48.37 &  96.04 \\
    \cellcolor{mygray-bg}{\ \ \emph{w/} {FL}}   & \cellcolor{mygray-bg}{\highest{40.14}} & \cellcolor{mygray-bg}{\highest{91.42}} \\
  \bottomrule

\end{tabular}
}

\vspace{0.3cm}

  \label{tab:spr}
  \vspace{-0.3cm}
\end{table*}
}

A CL method achieves a better stability-plasticity balance if the model trained by it demonstrates lower forgetting in old classes (indicating its stability) while effectively learning new classes (indicating plasticity). To monitor this balance, we define \textbf{Stability-Plasticity Ratio (SPR)} as:
\begin{equation}
    \label{eq:spr}
    \text{SPR} = \frac{\text{Forgetting on Old Classes} \,}{\text{Accuracy on New Classes} \,} \times 100 .
\end{equation}
The measurement of \textit{Forgetting on Old Classes} is given by Eq.~\eqref{eq:fg}, whereas \textit{Accuracy on New Classes} is the accuracy across all new classes in the last task. A lower SPR indicates a more favorable stability-plasticity tradeoff, allowing the model to avoid catastrophic forgetting of previous classes while effectively learning new ones. 
Table~\ref{tab:spr} presents the SPR results for CL representative baselines on Split~CIFAR-10 and Split~CIFAR-100 benchmarks, without and with FL augmentation. The results are averaged over several random seeds. If the condition $SPR_{\emph{w/}FL} < SPR_{\emph{w/o}FL}$ holds, the FL mechanism has contributed to improving the stability-plasticity balance. In Table~\ref{tab:spr}, adding FL to four CL methods across the two benchmarks consistently yields a smaller SPR value. 
\endgroup

\begingroup
\color{dp}
\subsection{Centered Kernel Alignment Analysis}
\label{sec:cka-results}

Centered Kernel Alignment (CKA)\cite{kornblith2019similarity} captures the similarity between two model representations on the same dataset at different layers. Consider a reference model $f_{r}$ that has seen the data from all tasks $\mathbb{T} = \{ \mathcal{T}_t\}_{t=1}^{T}$ simultaneously, so $f_{r}$ achieves the upper-bound accuracy on all tasks, including both old and new ones. The more the CL model can mimic $f_{r}$, the better it has been trained sequentially. We apply CKA to measure the similarity between the CL and reference models. An approach to analyzing stability-plasticity balance is to evaluate the model performance on old and new classes. To study this performance, we sequentially train the model on $T$ tasks both in CL and FL settings, respectively indicated by $f_{cl}$ and $f_{fl}$; we also divide the dataset into old and new classes, indicated by $\mathcal{D}_{\text{old}}$ and $\mathcal{D}_{\text{new}}$. Then we measure four CKA similarities between:
\begin{itemize}
    \item $f_{r}$ and $f_{cl}$ on $\mathcal{D}_{\text{old}}$  $\to$ $\text{CKA}_{f_{cl}, f_{r}}^{\text{old}}$,
    \item $f_{r}$ and $f_{fl}$ on $\mathcal{D}_{\text{old}}$  $\to$ $\text{CKA}_{f_{fl}, f_{r}}^{\text{old}}$,
    \item $f_{r}$ and $f_{cl}$ on $\mathcal{D}_{\text{new}}$  $\to$ $\text{CKA}_{f_{cl}, f_{r}}^{\text{new}}$,
    \item $f_{r}$ and $f_{fl}$ on $\mathcal{D}_{\text{new}}$  $\to$ $\text{CKA}_{f_{fl}, f_{r}}^{\text{new}}$.
\end{itemize}
Since early layers in DNNs are known to be highly transferable \cite{yosinski2014transferable}, deeper layers are more prone to changes in sequential training. Hence, we select representations from the final layers of the model for analysis. If the condition $\text{CKA}_{f_{cl}, f_{r}}^{\text{old}} < \text{CKA}_{f_{fl}, f_{r}}^{\text{old}}$ holds for selected layers, FL model $f_{fl}$ has resembled the reference model more closely than $f_{cl}$ for the old classes; otherwise, FL mechanism has not improved stability beyond that of the CL method. Moreover, the condition $\text{CKA}_{f_{cl}, f_{r}}^{\text{new}} < \text{CKA}_{f_{fl}, f_{r}}^{\text{new}}$ indicates that FL integration has improved the model performance in new classes.

By focusing on four layers from the last ResNet stage, we calculate $\text{CKA}_{f_{cl}, f_{r}}^{\text{new}}$ and $\text{CKA}_{f_{fl}, f_{r}}^{\text{new}}$ for iCaRL (Top Left Figure~\ref{fig4}). The findings reveal that $f_{fl}$ more effectively replicates the representations of the reference model in the selected layers compared to $f_{cl}$ for new classes. This pattern is also evident in CKAs for oEWC, before and after integrating FL (Bottom Left Figure~\ref{fig4}). A model's resemblance to the reference model in representation correlates with its ability to learn new classes effectively. We also compute $\text{CKA}_{f_{cl}, f_{r}}^{\text{old}}$ and $\text{CKA}_{f_{fl}, f_{r}}^{\text{old}}$, with results displayed in the right column of Figure~\ref{fig4}: iCaRL (Top Right) and oEWC (Bottom Right). The condition $\text{CKA}_{f_{cl}, f_{r}}^{\text{old}} < \text{CKA}_{f_{fl}, f_{r}}^{\text{old}}$ across all selected layers' representations for old classes suggests that $f_{fl}$ performs more closely to the reference model for these classes.

\begin{figure}[ht]
\centering
\includegraphics[width=0.7\columnwidth]{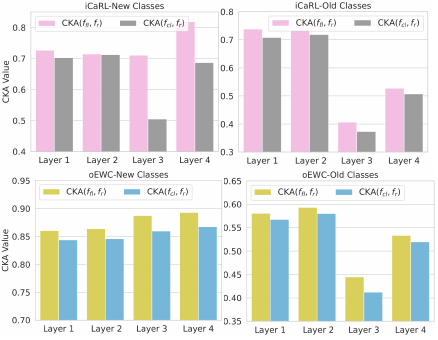} 
\caption{CKA Analysis on old and new test sets for oEWC and iCaRL in CL and FL settings}
\label{fig4}
\end{figure}

\endgroup

\section{Conclusion}

\begingroup
\color{dp}
We formulated the FL algorithm to smoothly integrate into prior methods from four main CL categories. Aiming to effectively guide the model to swiftly incorporate novel knowledge while actively retaining its old knowledge, the FL mechanism introduces a bidirectional form of regularization into existing methods. Our inspiration comes from biological studies, where memory refinement through integrating new information is vital in effective adaptation to new tasks. On the computational side, reverse distillation (where a student model can enhance its teacher) challenges the common belief in unidirectional knowledge transfer. To the best of our knowledge, our work is the first to demonstrate that bidirectional knowledge flow can benefit CL, showing that knowledge transfer in both directions is helpful.
FL operates in two phases: the first builds a knowledge base from new task data to promote model plasticity, and the second utilizes this knowledge base to counteract the unidirectional stability regularization in existing CL methods. Given the efficacy of our method across four main categories of CL methods, future research will explore FL's integration with emerging CL techniques, including parameter-efficient methods: prompt learning~(\eg \cite{wang2022learning}, \cite{smith2023coda}) and Low-rank adaptation~(\eg~\cite{liang2024inflora}, \cite{wang2023orthogonal}). By integrating FL into parameter-efficient methods, we can leverage large pre-trained models and move toward multi-modal continual learning. Additionally, we aim to extend the FL mechanism to more challenging CL settings, including online learning and task-agnostic scenarios. By pursuing these directions, we aim to advance the field of CL and contribute to developing more robust and adaptable learning systems.
\endgroup

\section*{Acknowledgments}
L.M. and P.M. gratefully acknowledge funding of the project by the CSIRO’s Data61 Science Digital and continued support from the CSIRO's Data61 Embodied AI Cluster. L.M. also acknowledges funding from the CSIRO’s Data61 Postgraduate Scholarship. M.H. gratefully acknowledges the support from the Australian Research Council (ARC), projects DP230101176 and DP250100262.

\section*{Declaration of generative AI and AI-assisted technologies}
During the preparation of this work, the author(s) used Grammarly and ChatGPT in order to improve language and readability. After using this tool/service, the author(s) reviewed and edited the content as needed and take(s) full responsibility for the content of the publication.
\appendix
\clearpage

\begingroup
\color{dp}
\section{Table of Notations}
\label{sec:not-table}
We provide a comprehensive list of all notations used across different paper sections in Table~\ref{tab:notation}.

\begin{table}[h]
\centering
\caption{Notation Table}
\resizebox{0.8\columnwidth}{!}{
\begin{tabular}{c l}
\hline
\textbf{Symbol} & \textbf{Definition} \\
\hline
& {\textbf{CL Notation}} \\
\hline
{$T$} & {Total number of tasks} \\
{$\mathcal{T}_t$} & {Task $t$} \\
{$\mathbb{T} = \{\mathcal{T}_i\}_{i=1}^T$} & {Sequence of tasks} \\
{$\data{t}$} & {Data distribution in task~$\mathcal{T}_t$} \\
{$\mathcal{X}_t$} & {Input distribution in task~$\mathcal{T}_t$} \\ 
{$\mathcal{Y}_t$} & {Output distribution in task~$\mathcal{T}_t$} \\ 
{$C_t$} & {Number of seen classes until task~$\mathcal{T}_t$}\\
{$\mpr{t}$} & {Parameter vector of model training on task~$\mathcal{T}_t$, $\mpr{t} \in \mathbb{R}^p$}  \\
{$\mdl{.}{t}{} =(g\circ h)(.;\mpr{t})$} & {Model training on task~$\mathcal{T}_t$, $\mpr{t} = \{\hpr{t}, \gpr{t}\}$}\\
{$\hpr{t}$} & {Parameter vector of task~$\mathcal{T}_t$'s feature extractor, $h(.;\hpr{t})$} \\
{$\gpr{t}$} & {Parameter vector of task~$\mathcal{T}_t$'s classifier, $g(.;\gpr{t})$} \\
{$h(.;\hpr{t}):\mathcal{X}_t \to \mathbb{R}^d$} & {
Task~$\mathcal{T}_t$'s feature extractor, $d$ is dimension of feature embedding space} \\
{$g(.;\gpr{t}):\mathbb{R}^d \to \mathbb{R}^{C_t}$} & {Task~$\mathcal{T}_t$'s classifier} \\
{$\emb{t} \in \mathbb{R}^d$} & {Task~$\mathcal{T}_t$'s feature embeddings} \\
{$\lgt{t}\in \mathbb{R}^c$} & {Task~$\mathcal{T}_t$'s classification logits} \\
{$\mpr{t}^*$} &  {Parameter vector of the model trained on task~$\mathcal{T}_t$, $\mpr{t}^* = \{\hpr{t}^*, \gpr{t}^*\}$} \\
{$\mdl{.}{t}{*}=(g\circ h)(.;\mpr{t}^*)$} & {Model trained on task~$\mathcal{T}_t$, is decomposed to feature extractor $h(.;\hpr{t}^*)$ and classifier $g(.;\gpr{t}^*)$} \\
{$\mathcal{M}_{t-1}$} & {Memory samples selected at conclusion of task~$\mathcal{T}_{t-1}$ for replay in task~$\mathcal{T}_t$} \\
{$\mdl{\vx}{}{}[i]$} & {The $i^{\text{th}}$ element of output vector, $i^{\text{th}}$ output of the logit} \\
{$\hat{\vy}(\vx;\tta)$} & {Model output after \textit{Softmax} layer} \\
\hline
& {\textbf{FL Notation}} \\
\hline
{$E_1$} & {Number of training epochs in \textit{Phase~1} of FL method} \\
{$E_2$} & {Number of training epochs in \textit{Phase~2} of FL method} \\
{$\mdl{.}{s}{}$} & {Stable model in task~$\mathcal{T}_t$, is model trained on task~$\mathcal{T}_{t-1}$: $\mdl{.}{s}{}=\mdl{.}{t-1}{*}$} \\
{$\mpr{s}, \hpr{s}, \gpr{s}$} & {Task~$\mathcal{T}_t$'s stable model parameters $\mpr{s}=\mpr{t-1}^*$, $\hpr{s}=\hpr{t-1}^*$, $\gpr{s}=\gpr{t-1}^*$} \\
{$\mdl{.}{p}{}$} & {Primary model in task~$\mathcal{T}_t$, is model training on task~$\mathcal{T}_t$ after \textit{Phase~1}: $\mdl{.}{p}{}=\mdl{.}{t}{}$} \\
{$\mpr{p}, \hpr{p}, \gpr{p}$} & {Task~$\mathcal{T}_t$'s primary parameters after \textit{Phase~1} $\mpr{p}=\mpr{t}$, $\hpr{p}=\hpr{t}$, $\gpr{p}=\gpr{t}$} \\
{$\mathcal{S}$} & {Task~$\mathcal{T}_t$'s stable knowledge kept from task~$\mathcal{T}_{t-1}$, defined per method~\textsection \ref{sec:skb}} \\
{$\mathcal{P}$} & {Plastic knowledge achieved in task~$\mathcal{T}_t$, defined per method~\textsection \ref{sec:pkb}} \\
{$\lgt{s}$} & {Stable logit in replay methods~\eqref{eq:skb-replay}, $\lgt{s}=f(\vx,\mpr{s})$} \\
{$\emb{s}$} & {Stable feature embedding in replay methods~\eqref{eq:skb-replay}, $\emb{s}=h(\vx,\hpr{s})$} \\
{$\mathbf{F}_t$} & {Fisher information matrix calculated in task~$\mathcal{T}_t$} \\
{$\hat{\mathbf{F}}_t$} & {Approximated Fisher information matrix in task~$\mathcal{T}_t$~\eqref{eq:fisher2}} \\
{$\mathbf{F}_s$} & {Stable Fisher information matrix in task~$\mathcal{T}_t$~\eqref{eq:skb-reg}, $\mathbf{F}_s = \hat{\mathbf{F}}_{t-1}$} \\
{$h(.;\hpr{s})$} & {Stable feature extractor in task~$\mathcal{T}_t$~\eqref{eq:skb-dyn}, $h(.;\hpr{s})=h(.;\hpr{t-1}^*)$} \\
{$m(.;\nhpr{t})$} & {New module added to stable feature extractor in task~$\mathcal{T}_t$~\eqref{eq:dyn-expansion} in dynamic architecture methods} \\
{$\lgt{p}$} & {Plastic logit in replay methods~\eqref{eq:pkb-replay}, $\lgt{p}=f(\vx,\mpr{p})$} \\
{$\emb{p}$} & {Plastic embedding in replay methods~\eqref{eq:pkb-replay}, $\emb{p}=h(\vx,\hpr{p})$} \\
{$\mathbf{F}_p$} & {Plastic Fisher information matrix in task~$\mathcal{T}_t$~\eqref{eq:pkb-reg}, $\mathbf{F}_p = \mathbf{F}_{t}$ after \textit{Phase~1}} \\
{$m(.;\nhpr{p})$} & {Plastic module in dynamic architecture methods~\eqref{eq:pkb-dyn}, $m(.;\nhpr{p}) = m(.;\nhpr{t})$ after \textit{Phase~1}} \\
{$\mathcal{L}_c(\tta)$} & {Task-specific loss (\eg classification loss)} \\
{$\mathcal{L}_s(\tta)$} & {Stability loss, defined per CL method~\textsection~\ref{sec:phase1}} \\
{$\mathcal{L}_p(\tta)$} & {Plasticity loss, defined per CL method~\textsection~\ref{sec:phase2}} \\
{$\alpha_s$} & {Stability loss scaler} \\
{$\alpha_p$} & {Plasticity loss scaler} \\
\hline
\end{tabular}}
\label{tab:notation}
\end{table}


\section{Cosine Embedding Loss}
\label{sec:general-dist}
Consider the cosine embedding loss in LUCIR~\cite{hou2019learning}:
\begin{align*}
    \mathcal{L}_s (\tta)=\mathbb{E}_{\vx  \sim \mathcal{X}_t} \Big[ 1- \langle \bar{f}(\vx;\tta) , \bar{f}(\vx;\tta_s) \rangle \Big].
\end{align*}
Let $\bar{f} = f / \lVert f \rVert_2$ represent the $\ell_2$-normalized vector of the output vectors and $\langle \cdot, \cdot\rangle$ is the inner product between two vectors. The dot product between two outputs (\ie $\bar{f}(\vx;\mpr{})$ and $\bar{f}(\vx;\mpr{s}))$ can be rewritten in terms of the squared difference between them. Use the identity:
\begin{equation}
    \left\| \bar{f}(\vx;\tta) - \bar{f}(\vx;\tta_s) \right\|^2 =  \left\| \bar{f}(\vx;\tta) \right\|^2 +  \left\| \bar{f}(\vx;\tta_s) \right\|^2 - 2 \langle \bar{f}(\vx;\tta) , \bar{f}(\vx;\tta_s) \rangle.
\end{equation}
Since embeddings are normalized we have $\lVert \bar{f}(\vx;\tta) \rVert = 1$, $\lVert \bar{f}(\vx;\tta_s) \rVert = 1$, and:
\begin{equation}
\label{eq:dot-equal}
    \left\| \bar{f}(\vx;\tta) - \bar{f}(\vx;\tta_s) \right\|^2 =  2 - 2 \langle \bar{f}(\vx;\tta) , \bar{f}(\vx;\tta_s) \rangle \quad \implies \quad \langle \bar{f}(\vx;\tta) , \bar{f}(\vx;\tta_s) \rangle= 1 - \frac{1}{2} \left\| \bar{f}(\vx;\tta) - \bar{f}(\vx;\tta_s) \right\|^2
\end{equation}
Substitute \eqref{eq:dot-equal} in the stability loss, we have the general form:
\begin{equation}
\label{eq:distill-general}
    \mathcal{L}_s(\tta) =  \mathbb{E}_{\vx \sim \mathcal{X}_t} \Big[ \frac{1}{2} \left\| \bar{f}(\vx;\tta) - \bar{f}(\vx;\tta_s) \right\|^2 \Big]\;.
\end{equation}

\section{Proofs}
\label{sec:proofs}

We provide the proof of all theorems stated in \textsection\ref{sec:analysis} here.

\subsection{Gradient Decomposition for Distillation Methods}
\label{subsec:p-distillation}

\begin{proof}
In \uline{CL case}, suppose the loss function $\mathcal{L}_{\text{CL}}(\mpr{})$ for distillation methods be of form:
\begin{equation}
\label{eq:l-lucir-2}
    \mathcal{L}_{\text{CL}}(\tta) =  \underbrace{\mathcal{L}_{c}(\tta)}_{\text{Task-specific Loss}} + \alpha_s \underbrace{\mathbb{E}_{\vx \sim \mathcal{X}_t} \Big[ \frac{1}{2} \left\| f(\vx;\tta) - f(\vx;\tta_s) \right\|^2 \Big]}_{\text{Stability Loss}};
\end{equation}
then take the gradient \textit{w.r.t} $\tta$ at a sample $(\vx,\vy) \sim \mathcal{D}_t$:
\begin{equation}
\label{eq:p-dist}
    \nabla_{\mpr{}}\mathcal{L}_{\text{CL}}(\tta) =  \underbrace{\nabla_{\mpr{}}\mathcal{L}_{c}(\tta)}_{\text{Task-specific Gradient}} +  \alpha_s \underbrace{ \nabla_{\tta} f(\vx;\tta)^\top \left( f(\vx;\tta) - f(\vx;\tta_s) \right) }_{\text{Stability Gradient}}.
\end{equation}
Eq.~\eqref{eq:p-dist} shows the gradient in the CL setting, including task-specific and stability terms.
\\

In \uline{FL case}, we consider $\mathcal{L}_{\text{FL}}(\tta)$ with stability~\eqref{eq:ls-distil} and plasticity~\eqref{eq:lp-distil} distillation terms:
\begin{equation}
\label{eq:l-lucir-f-2}
    \mathcal{L}_{\text{FL}}(\tta) =  \underbrace{\mathcal{L}_{c}(\tta)}_{\text{Task-specific Loss}} + \alpha_s \underbrace{\mathbb{E}_{\vx \sim \mathcal{X}_t} \Big[ \frac{1}{2} \left\| f(\vx;\tta) - f(\vx;\tta_s) \right\|^2 \Big]}_{\text{Stability Loss}}+ \alpha_p \underbrace{\mathbb{E}_{\vx \sim \mathcal{X}_t} \Big[ \frac{1}{2} \left\| f(\vx;\tta) - f(\vx;\tta_p) \right\|^2 \Big]}_{\text{Plasticity Loss}}.
\end{equation}
We take the gradient of $\mathcal{L}_{\text{FL}}(\tta)$~\eqref{eq:l-lucir-f-2} w.r.t $\tta$ at sample $(\vx,\vy) \sim \mathcal{D}_t$:
\begin{equation}
\label{eq:dl-lucir-f}
    \nabla_{\mpr{}}\mathcal{L}_{\text{FL}}(\tta) =  \underbrace{\nabla_{\mpr{}}\mathcal{L}_{c}(\tta)}_{\text{Task-specific Gradient}} +  \alpha_s \underbrace{ \nabla_{\tta} f(\vx;\tta)^\top \left( f(\vx;\tta) - f(\vx;\tta_s) \right)   }_{\text{Stability Gradient}} + \alpha_p \underbrace{ \nabla_{\tta} f(\vx;\tta)^\top \left( f(\vx;\tta) - f(\vx;\tta_p) \right)   }_{\text{Plasticity Gradient}};
\end{equation}
and rewrite it as:
\begin{equation}
\label{eq:p-fdist}
    \nabla_{\tta} \mathcal{L}_{\text{FL}}(\tta) =  \underbrace{\nabla_{\tta} \mathcal{L}_{c}(\tta)}_{\text{Task-specific Gradient}} + \left(\alpha_s+\alpha_p\right) \underbrace{\nabla_{\tta} f(\vx;\tta)^\top \left( f(\vx;\tta) - \frac{\alpha_s f(\vx;\tta_s) + \alpha_p f(\vx;\tta_p)}{\alpha_s+\alpha_p} \right)  }_{\text{Gradient Interpolation}}. 
\end{equation}

Eq.~\eqref{eq:p-fdist} shows the gradient in the FL setting, including task-specific and interpolation terms.
\end{proof}
\subsection{Gradient Decomposition for Replay Methods}
\label{subsec:p-replay}

\begin{proof}
In \uline{CL case}, we have $\mathcal{L}_{\text{CL}}(\tta)$ in the following form: 
\begin{equation}
\label{eq:l-der}
    \mathcal{L}_{\text{CL}}(\tta) =  \underbrace{\mathcal{L}_{c}(\tta)}_{\text{Task-specific Loss}} + \alpha_s  \underbrace{\mathbb{E}_{(\vx,\lgt{s}) \sim \mathcal{S}} \Big[ \frac{1}{2} \big \lVert \mdl{\vx}{}{} - \lgt{s} \big \rVert^2 \Big]}_{\text{Stability Loss}},
\end{equation}
which takes $\mathcal{S}$ from SKB~\eqref{eq:skb-replay}. We compute the gradient of~\eqref{eq:l-der} w.r.t $\tta$, using one sample from memory $(\vx, \vy ) \sim \mathcal{M}_{t-1}$ and its corresponding logit $\lgt{s}$~\eqref{eq:skb-replay}:
\begin{equation}
\label{eq:p-replay}
    \nabla_{\tta} \mathcal{L}_{\text{CL}}(\tta) =  \underbrace{\nabla_{\tta} \mathcal{L}_{c}(\tta)}_{\text{Task-specific Gradient}} +  \alpha_s \underbrace{ \nabla_{\tta} f(\vx;\tta)^\top \left( f(\vx;\tta) - \lgt{s} \right) }_{\text{Stability Gradient}}.
\end{equation}
Eq.~\eqref{eq:p-replay} defines the gradient form in the CL setting.
\\

In \uline{FL case}, with $\mathcal{L}_{\text{FL}}(\tta)$ in the form of:
\begin{equation}
\label{eq:l-der-f}
    \mathcal{L}_{\text{FL}}(\tta) =  \underbrace{\mathcal{L}_{c}(\tta)}_{\text{Task-specific Loss}} + \alpha_s  \underbrace{\mathbb{E}_{(\vx,\lgt{s}) \sim \mathcal{S}} \Big[ \frac{1}{2} \big \lVert \mdl{\vx}{}{} - \lgt{s} \big \rVert^2 \Big]}_{\text{Stability Loss}} + \alpha_p  \underbrace{\mathbb{E}_{(\vx,\lgt{p}) \sim (\mathcal{M}_{t-1},\mathcal{P})} \Big[ \frac{1}{2} \big \lVert \mdl{\vx}{}{} - \lgt{p} \big \rVert^2 \Big]}_{\text{Plasticity Loss}};
\end{equation}
we compute the gradient at the same memory sample $(\vx,\vy ) \sim \mathcal{M}_{t-1}$, using the corresponding stable logit~$\lgt{s}$~\eqref{eq:skb-replay} and primary logit $\lgt{p}$~\eqref{eq:pkb-replay}:
\begin{equation}
\label{eq:dl-der-f}
    \nabla_{\tta} \mathcal{L}_{\text{FL}}(\tta) =  \underbrace{\nabla_{\tta} \mathcal{L}_{c}(\tta)}_{\text{Task-specific Gradient}} +  \alpha_s \underbrace{ \nabla_{\tta} f(\vx;\tta)^\top \left( f(\vx;\tta) - \lgt{s} \right) }_{\text{Stability Gradient}} + \alpha_p \underbrace{ \nabla_{\tta} f(\vx;\tta)^\top \left( f(\vx;\tta) - \lgt{p} \right) }_{\text{Plasticity Gradient}}.
\end{equation}
We can rewrite \eqref{eq:dl-der-f}, to have:
\begin{equation}
\label{eq:p-freplay}
    \nabla_{\tta} \mathcal{L}_{\text{FL}}(\tta) =  \underbrace{\nabla_{\tta} \mathcal{L}_{c}(\tta)}_{\text{Task-specific Gradient}} +  \left(\alpha_s+\alpha_p\right) \underbrace{ \nabla_{\tta} f(\vx;\tta)^\top \left( f(\vx;\tta) - \frac{\alpha_s \lgt{s} + \alpha_p \lgt{p}}{\alpha_s+\alpha_p} \right)  }_{\text{Gradient Interpolation}}.
\end{equation}
\end{proof}
Eq.~\eqref{eq:p-freplay} defines the gradient form after integration with FL.
\subsection{Gradient Decomposition for Regularization Methods}
\label{subsec:p-regularize}

\begin{proof}
In \uline{CL case}, consider the loss function~$\mathcal{L}_{\text{CL}}(\tta)$ and its stability loss component~\eqref{eq:ls-reg} for regularization methods:
\begin{equation}
\label{eq:l-ewc}
    \mathcal{L}_{\text{CL}}(\tta) =  \underbrace{\mathcal{L}_{c}(\tta)}_{\text{Task-specific Loss}} + \frac{\alpha_s}{2}\underbrace{\left(\tta - \tta_s \right)^\top \mathbf{F}_s \left(\tta - \tta_s \right)}_{\text{Stability Loss}}.
\end{equation}
We take the gradient of $\mathcal{L}_{\text{CL}}(\tta)$ with respect to~$\tta$:
\begin{equation}
\label{eq:dl-ewc}
    \nabla_{\mpr{}} \mathcal{L}_{\text{CL}}(\tta) =  \underbrace{\nabla_{\mpr{}} \mathcal{L}_{c}(\tta)}_{\text{Task-specific Gradient}} + \underbrace{\alpha_s 
 \mathbf{F}_{s} \left(\mpr{} -\mpr{s} \right)}_{\text{Stability Gradient}}. 
\end{equation}
Eq.~\eqref{eq:dl-ewc} presents the gradient in the CL setting, where $\mathbf{F}_{s}$ represents the stable FIM kept in SKB~\eqref{eq:skb-reg}.
\\

In \uline{FL case}, having stability~\eqref{eq:ls-reg} and plasticity~\eqref{eq:lp-reg} terms in the loss function $\mathcal{L}_{\text{FL}}(\tta)$:
\begin{equation}
\label{eq:l-ewc-f}
    \mathcal{L}_{\text{FL}}(\tta) =  \underbrace{\mathcal{L}_{c}(\tta)}_{\text{Task-specific Loss}} + \frac{\alpha_s}{2} \underbrace{\left(\tta - \tta_s \right)^\top \mathbf{F}_s \left(\tta - \tta_s \right)}_{\text{Stability Loss}}+\frac{\alpha_p}{2}  \underbrace{\left(\tta - \tta_p \right)^\top \mathbf{F}_p \left(\tta - \tta_p \right)}_{\text{Plasticity Loss}};
\end{equation}
we take the gradient with respect to $\mpr{}$:
\begin{equation}
\label{eq:dl-ewc-f}
    \nabla_{\mpr{}} \mathcal{L}_{\text{FL}}(\tta) =  \underbrace{\nabla_{\mpr{}} \mathcal{L}_{c}(\tta)}_{\text{Task-specific Gradient}} + \underbrace{\alpha_s \mathbf{F}_{s} \left(\mpr{} - \mpr{s} \right) + \alpha_p \mathbf{F}_{p} \left(\mpr{} - \mpr{p} \right)}_{\text{Gradient Interpolation}}.
\end{equation}
\end{proof}
Eq.~\eqref{eq:dl-ewc-f} shows the gradient in the FL setting, where $\tta_s / \mathbf{F}_{s}$ and $\tta_p /\mathbf{F}_{p}$ represent the stable and primary Parameter/FIM stored in SKB~\eqref{eq:skb-reg} and PKB~\eqref{eq:pkb-reg} respectively.
\subsection{Gradient Decomposition for Dynamic Architecture Methods}
\label{subsec:p-dynamic} 
\begin{proof}
Let's denote the model output after the softmax layer as:
\begin{equation}
\label{eq:y-hat}
     \hat{\vy}(\vx;\mpr{}) =  \frac{\exp{\big(\mdl{\vx}{}{}/\tau \big)}}{\sum_{k=1}^{C_t}{\exp{\big(\mdl{\vx}{}{}[k]/\tau\big)}}};
\end{equation}
where $\tau$ is the temperature parameter, and $\mdl{\vx}{}{}[k]$ is the $k^{\text{th}}$ logit of the model output. 

In \uline{CL case}, using the KL stability loss~\eqref{eq:ls-dyn} to align the probability distribution modeled by $\mdl{\cdot}{}{}$ with distribution modeled by $\mdl{\cdot}{s}{}$, the loss function is expressed as:
\begin{equation}
    \label{eq:l-foster}
    \mathcal{L}_{\text{CL}}(\tta) =  \underbrace{\mathcal{L}_{c}(\tta)}_{\text{Task-specific Loss}} - \alpha_s \underbrace{ \mathbb{E}_{\vx \sim \mathcal{X}_t} \bigg(  \hat{\vy}(\vx;\mpr{s})^\top \log \hat{\vy}(\vx;\mpr{}) \bigg)}_{\text{Stability Loss}}.
\end{equation}
We compute the gradient of the CL loss function with respect to $\mpr{}$ at sample $(\vx , \vy) \sim \mathcal{D}_t$ as follows~\cite{bishop2023deep}:
\begin{equation}
    \label{eq:dl-foster}
    \nabla_{\mpr{}}\mathcal{L}_{\text{CL}}(\tta) =  \underbrace{\nabla_{\mpr{}}\mathcal{L}_{c}(\tta)}_{\text{Task-specific Gradient}} + \frac{\alpha_s}{\tau} \underbrace{ \nabla_{\mpr{}} \mdl{\vx}{}{}^\top \big( \hat{\vy}(\vx;\mpr{}) - \hat{\vy}(\vx;\mpr{s}) \big) }_{\text{Stability Gradient}}.
\end{equation}
Eq.~\eqref{eq:dl-foster} presents the gradient in the CL setting, where $\hat{\vy}(\vx;\mpr{s})$ represents the probability distribution parameterized by the model with stable feature extractor~\eqref{eq:skb-dyn}.
\\

In \uline{FL case}, the objective function is constructed using stability~\eqref{eq:ls-dyn}
and plasticity~\eqref{eq:lp-dyn} KL losses:
\begin{equation}
\label{eq:l-foster-f}
\mathcal{L}_{\text{FL}}(\tta) =  \underbrace{\mathcal{L}_{c}(\tta)}_{\text{Task-specific Loss}} - \alpha_s \underbrace{ \mathbb{E}_{\vx \sim \mathcal{X}_t} \bigg(  \hat{\vy}(\vx;\mpr{s})^\top \log \hat{\vy}(\vx;\mpr{}) \bigg)}_{\text{Stability Loss}} - \alpha_p \underbrace{\mathbb{E}_{\vx \sim \mathcal{X}_t} \bigg(  \hat{\vy}(\vx;\mpr{p})^\top \log \hat{\vy}(\vx;\mpr{}) \bigg) }_{\text{Plasticity Loss}}.
\end{equation}
With the same notation as the previous scenario, we write the gradient of~$\mathcal{L}_{\text{FL}}(\tta)$:
\begin{align}
    \label{eq:dl-foster-f}
    \nabla_{\mpr{}}\mathcal{L}_{\text{FL}}(\tta) =  \underbrace{\nabla_{\mpr{}}\mathcal{L}_{c}(\tta)}_{\text{Task-specific Gradient}} &+ \frac{\alpha_s}{\tau} \underbrace{ \nabla_{\mpr{}} \mdl{\vx}{}{}^\top \big( \hat{\vy}(\vx;\mpr{}) - \hat{\vy}(\vx;\mpr{s}) \big) }_{\text{Stability Gradient}} \notag \\
    & + \frac{\alpha_p}{\tau}  \underbrace{ \nabla_{\mpr{}} \mdl{\vx}{}{}^\top \big( \hat{\vy}(\vx;\mpr{}) - \hat{\vy}(\vx;\mpr{p}) \big) }_{\text{Plasticity Gradient}};
\end{align}
then rewrite the gradient as:
\begin{equation}
\label{eq:dl-f-foster}
    \nabla_{\mpr{}}\mathcal{L}_{\text{FL}}(\tta) =  \underbrace{\nabla_{\mpr{}}\mathcal{L}_{c}(\tta)}_{\text{Task-specific Gradient}} + \frac{\alpha_s + \alpha_p}{\tau}  \underbrace{ \nabla_{\mpr{}} \mdl{\vx}{}{}^\top \bigg( \hat{\vy}(\vx;\mpr{}) - \frac{\alpha_s \, \hat{\vy}(\vx;\mpr{s})+ \alpha_p \, \hat{\vy}(\vx;\mpr{p})}{\alpha_s + \alpha_p} \bigg)  }_{\text{Gradient Interpolation}}.
\end{equation}
Eq.~\eqref{eq:dl-f-foster} presents the gradient in the FL setting, where $\hat{\vy}(\vx;\mpr{s})$ and $\hat{\vy}(\vx;\mpr{p})$ represent the probability distributions parameterized by the stable model with stable feature extractor~\eqref{eq:skb-dyn} and the primary model incorporating primary new module~\eqref{eq:pkb-dyn}.
\end{proof}

\subsection{SGD Trajectories in Parameter Regularization Methods}
\label{subsec:update-regularize-p}
Consider gradient expression from ~\eqref{eqn:grad_cl_decomp_reg_inf} for the loss function in the \uline{CL case} under regularization methods:
\begin{align*}
     \nabla_{\tta} \mathcal{L}_{\text{CL}}(\tta) =  \underbrace{\nabla_{\tta} \mathcal{L}_{c}(\tta)}_{\text{Task-specific Gradient}} +  
      \underbrace{ \alpha_s \mathbf{F}_s \big(\tta - \tta_s  \big)}_{\text{Stability Gradient}}. 
\end{align*}
Substituting it in the Stochastic Gradient Descent (SGD) update rule yields:
\begin{equation}
\label{eq:e1}
    \tta^{(k+1)} = \tta^{(k)} - \eta \Big( \nabla_{\tta} \mathcal{L}_{c}(\tta)\bigg|_{\tta=\tta^{(k)}} \Big) -  \eta \alpha_s \mathbf{F}_s \big( \tta^{(k)} - \tta_s \big);
\end{equation}
and we rewrite it in the following form:
\begin{equation}
\label{eq:e1-1}
    \tta^{(k+1)} = \big(\mathrm{I} - \eta \alpha_s \mathbf{F}_s \big) \tta^{(k)} - \eta \Big( \nabla_{\tta} \mathcal{L}_{c}(\tta)\bigg|_{\tta=\tta^{(k)}} \Big) +  \eta \alpha_s \mathbf{F}_s \tta_s ;
\end{equation}
where $\tta^{(k)}$ denotes the parameter value at training iteration~$k$. Starting from $\tta^{(0)}$, we derive a recursive formula to express $\tta^{(k)}$ based on its initial value: 
\begin{align}
\tta^{(1)} =& \underbrace{\big(\mathrm{I} - \eta \alpha_s \mathbf{F}_s \big) \tta^{(0)} - \eta \Big( \nabla_{\tta} \mathcal{L}_{c}(\tta)\bigg|_{\tta=\tta^{(0)}} \Big) +  \eta \alpha_s \mathbf{F}_s \tta_s}_{\text{A}} , \notag\\
\tta^{(2)} =& \big(\mathrm{I} - \eta \alpha_s \mathbf{F}_s \big) \underbrace{\tta^{(1)}}_{\text{A}} - \eta \Big( \nabla_{\tta} \mathcal{L}_{c}(\tta)\bigg|_{\tta=\tta^{(1)}} \Big) +  \eta \alpha_s \mathbf{F}_s \tta_s  \notag \\
=& \big(\mathrm{I} - \eta \alpha_s \mathbf{F}_s \big)^2 \tta^{(0)} - \eta \sum_{j=0}^1\big(\mathrm{I} - \eta \alpha_s \mathbf{F}_s \big)^{1-j}  \Big( \nabla_{\tta} \mathcal{L}_{c}(\tta)\bigg|_{\tta=\tta^{(j)}} \Big) + \sum_{j=0}^1 \big(\mathrm{I} - \eta \alpha_s  \mathbf{F}_s \big)^{1-j} \big( \eta \alpha_s \mathbf{F}_s \tta_s \big)^j  \notag \\
& \vdots \notag \\ 
\tta^{(k)} =& \big(\mathrm{I} - \eta \alpha_s \mathbf{F}_s \big)^k \tta^{(0)} - \eta \sum_{j=0}^{k-1} \big(\mathrm{I} - \eta \alpha_s \mathbf{F}_s \big)^{k-1-j}  \Big( \nabla_{\tta} \mathcal{L}_{c}(\tta)\bigg|_{\tta=\tta^{(j)}} \Big) + \sum_{j=0}^{k-1} \big(\mathrm{I} - \eta \alpha_s  \mathbf{F}_s \big)^{k-1-j} \big(\eta \alpha_s \mathbf{F}_s \tta_s \big)^{j} \label{eq:recursive-cl}
\end{align}

Let denote $\Gamma_s = \eta \alpha_s \mathbf{F}_s$ in the Eq.~\eqref{eq:recursive-cl} to obtain:
\begin{equation}
\tta^{(k)} = \big(\mathrm{I} - \Gamma_s \big)^k \tta^{(0)} - \eta \sum_{j=0}^{k-1} \big(\mathrm{I} - \Gamma_s \big)^{k-1-j}  \Big( \nabla_{\tta} \mathcal{L}_{c}(\tta)\bigg|_{\tta=\tta^{(j)}} \Big) + \sum_{j=0}^{k-1} \big(\mathrm{I} - \Gamma_s \big)^{k-1-j} \big(\Gamma_s \tta_s \big)^{j} \label{eq:recursive-cl-2}.
\end{equation}
Eq.~\eqref{eq:recursive-cl-2} shows the SGD trajectory in the CL setting.
\\

In \uline{FL case}, the gradient is given by~\eqref{eqn:grad_fl_decomp_reg_inf}:
\begin{align*}
    \nabla_{\tta} \mathcal{L}_{\text{FL}}(\tta) =  \underbrace{\nabla_{\tta} \mathcal{L}_{c}(\tta)}_{\text{Task-specific Gradient}} +  
    \underbrace{\alpha_s \mathbf{F}_s\big(\tta - \tta_s  \big) +   \alpha_p \mathbf{F}_p\big(\tta - \tta_p  \big)  }_{\text{Gradient Interpolation}}.
\end{align*}
Substituting this into the SGD update rule results in:
\begin{equation}
\label{eq:e3}
    \tta^{(k+1)} = \big(I-  \eta \alpha_s \mathbf{F}_s -  \eta \alpha_p \mathbf{F}_p\big) \tta^{(k)} - \eta \Big( \nabla_{\tta} \mathcal{L}_{c}(\tta)\bigg|_{\tta=\tta^{(k)}} \Big) +  \eta \alpha_s \mathbf{F}_s +  \eta \alpha_p \mathbf{F}_p .
\end{equation}
From Eq.~\eqref{eq:e3}, we derive a recursive formula for $ \tta^{(k)}$:
\begin{align}
\tta^{(1)} =& \underbrace{\big(\mathrm{I} - \eta \alpha_s \mathbf{F}_s - \eta \alpha_p \mathbf{F}_p \big) \tta^{(0)} - \eta \Big( \nabla_{\tta} \mathcal{L}_{c}(\tta)\bigg|_{\tta=\tta^{(0)}} \Big) +  \eta \big( \alpha_s \mathbf{F}_s \tta_s + \alpha_p \mathbf{F}_p \tta_p \big) }_{\text{A}} , \notag\\
\tta^{(2)} =& \big(\mathrm{I} - \eta \alpha_s \mathbf{F}_s - \eta \alpha_p \mathbf{F}_p \big) \underbrace{\tta^{(1)}}_{\text{A}} - \eta \Big( \nabla_{\tta} \mathcal{L}_{c}(\tta)\bigg|_{\tta=\tta^{(1)}} \Big) +  \eta \big( \alpha_s \mathbf{F}_s \tta_s + \alpha_p \mathbf{F}_p \tta_p \big)  \notag \\
=& \big(\mathrm{I} - \eta \alpha_s \mathbf{F}_s - \eta \alpha_p \mathbf{F}_p \big)^2 \tta^{(0)} - \eta \sum_{j=0}^1\big(\mathrm{I} - \eta \alpha_s \mathbf{F}_s - \eta \alpha_p \mathbf{F}_p \big)^{1-j}  \Big( \nabla_{\tta} \mathcal{L}_{c}(\tta)\bigg|_{\tta=\tta^{(j)}} \Big) + \notag \\
&\sum_{j=0}^1 \big(\mathrm{I} - \eta \alpha_s \mathbf{F}_s - \eta \alpha_p \mathbf{F}_p \big)^{1-j} \big( \eta \alpha_s \mathbf{F}_s \tta_s + \eta \alpha_p \mathbf{F}_p \tta_p \big)^j  \notag \\
& \vdots \notag \\ 
\tta^{(k)} =& \big(\mathrm{I} - \eta \alpha_s \mathbf{F}_s - \eta \alpha_p \mathbf{F}_p \big)^k \tta^{(0)} - \eta \sum_{j=0}^{k-1} \big(\mathrm{I} - \eta \alpha_s \mathbf{F}_s - \eta \alpha_p \mathbf{F}_p \big)^{k-1-j}  \Big( \nabla_{\tta} \mathcal{L}_{c}(\tta)\bigg|_{\tta=\tta^{(j)}} \Big) + \notag \\
&\sum_{j=0}^{k-1} \big(\mathrm{I} - \eta \alpha_s \mathbf{F}_s - \eta \alpha_p \mathbf{F}_p \big)^{k-1-j} \big( \eta \alpha_s \mathbf{F}_s \tta_s + \eta \alpha_p \mathbf{F}_p \tta_p \big)^{j} \label{eq:recursive-fl}
\end{align}
Define $\Gamma_s = \eta \alpha_s \mathbf{F}_s$ and $\Gamma_p = \eta \alpha_p \mathbf{F}_p$ in the Eq.~\eqref{eq:recursive-fl} to obtain:
\begin{align}
  \tta^{(k)} =& \big(\mathrm{I} - \Gamma_s - \Gamma_p \big)^k \tta^{(0)} - \eta \sum_{j=0}^{k-1} \big(\mathrm{I} - \Gamma_s - \Gamma_p \big)^{k-1-j}  \Big( \nabla_{\tta} \mathcal{L}_{c}(\tta)\bigg|_{\tta=\tta^{(j)}} \Big) + \notag \\
&\sum_{j=0}^{k-1} \big(\mathrm{I} - \Gamma_s - \Gamma_p \big)^{k-1-j} \big( \Gamma_s \tta_s + \Gamma_p \tta_p \big)^{j} \label{eq:recursive-fl-1}  
\end{align}
Eq.~\eqref{eq:recursive-fl-1} characterizes the SGD trajectory in the FL setting. 

\endgroup

\newpage
\bibliographystyle{elsarticle-num}

\newpage

\end{document}